%% file: main.tex
\documentclass[final]{cvpr}

\usepackage{times}
\usepackage{epsfig}
\usepackage{graphicx}
\usepackage{amsmath}
\usepackage{amssymb}
\usepackage{subfig}

\usepackage{booktabs}       
\usepackage{amsfonts}       
\usepackage{nicefrac}       
\usepackage{microtype}      
\usepackage{wrapfig}
\usepackage{xcolor}

\usepackage{caption}
\usepackage[pagebackref=true,breaklinks=true,colorlinks,bookmarks=false]{hyperref}

\def\cvprPaperID{4779} 
\def\confYear{CVPR 2021}

\begin{document}

\title{Self-Supervised Equivariant Scene Synthesis from Video}

\author{%
  Cinjon Resnick \\
    NYU \\
    {\tt\small cinjon@nyu.edu}
    \and
  Or Litany \\
  NVidia \\
  \and 
  Cosmas Heiß \\
  Technical University of Berlin \\
  \and
  Hugo Larochelle \\
  Google \\
  \and
  Joan Bruna \\
  NYU \\
  \and
  Kyunghyun Cho \\
  NYU 
}

\maketitle

\input{core/0_abstract}

\input{core/1_introduction}

\input{core/3_background}

\input{core/4_method}

\input{core/5_experiments}
\input{core/2_related_work}
\input{core/7_conclusion}

\input{core/8_broader_impact}

\clearpage
{\small
\bibliographystyle{ieee_fullname}
\bibliography{refs}
}


\clearpage
\newpage
\appendix

\input{appendix}

\end{document}

%% file: core/0_abstract.tex
\begin{abstract}

We propose a self-supervised framework to learn scene representations from video that are automatically delineated into background, characters, and their animations. Our method capitalizes on moving characters being equivariant with respect to their transformation across frames and the background being constant with respect to that same transformation. After training, we can manipulate image encodings in real time to create unseen combinations of the delineated components. As far as we know, we are the first method to perform unsupervised extraction and synthesis of interpretable background, character, and animation. We demonstrate results on three datasets: Moving MNIST with backgrounds, 2D video game sprites, and Fashion Modeling.

\end{abstract}

%% file: core/1_introduction.tex
\section{Introduction}
\label{sec:intro}


Learning manipulable representations of dynamic scenes is a challenging task. Ideally, we would give our model a video and receive an inverse rendering of coherent moving objects (`characters') along with a set of static backgrounds through which those characters move.

A motivating example is creating a stop motion video in which each character is painstakingly moved across static backgrounds to form the frames. Our goal is to have a model train on any number of videos, each of which have an independent set of characters and background, and then be able to synthesize new scenes mixing and matching (possibly unseen) characters and the backgrounds, thereby expediting the process of creating stop motion videos. This long standing problem with no good solution (see Section~\ref{sec:related-work}) can be thought of as unsupervised separation and synthesis of background and foreground, and a strong solution would address other applications as well, such as video compositing. Current unsupervised methods performing synthesis are limited in that they either cannot handle backgrounds~\cite{NIPS2015_5845,visualdynamics16,Kosiorek2018sqair}, cannot synthesize the foreground and background independently~\cite{DBLP:journals/corr/VillegasYHLL17,Tulyakov_2018_CVPR,siarohin2020order}, or cannot handle animation~\cite{DBLP:journals/corr/VondrickPT16}. 

\begin{figure}[t]
    \centering
    \includegraphics[width=7cm]{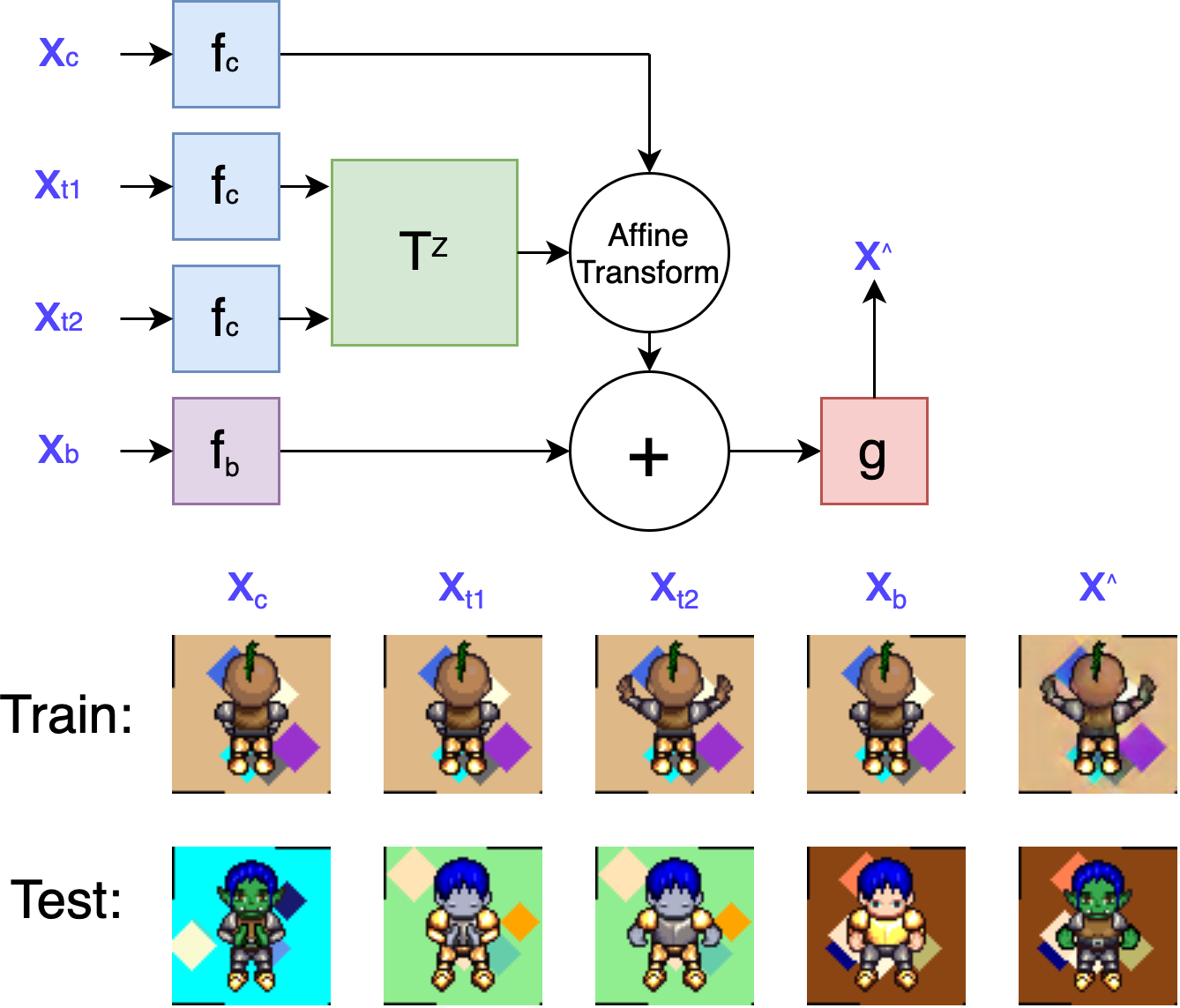}
    \caption{\textbf{Approach}: We use neural networks to parameterize $f_c$, $f_b$, $g$, and $T^{Z}$. With $f_c$, we encode the image to be transformed, $x_{\mathrm{c}}$, as well as the two images, $x_{\mathrm{t1}}$ and $x_{\mathrm{t2}}$, whose transformation we impose on $x_{\mathrm{c}}$. The output of $T^Z$ is an affine matrix that we apply to $f_c(x_{\mathrm{c}})$. That result is added to the background encoding $f_b(x_{\mathrm{b}})$ and then decoded to yield $\hat{x} = g(T^{Z}(f_c(x_{\mathrm{t1}}), f_c(x_{\mathrm{t2}})) \circ f_c(x_{\mathrm{c}}) + f_b(x_{\mathrm{b}}))$. For training, we require only two distinct images, $(x_i, x_j)$, from the same video, where $x_i = x_{\mathrm{c}} = x_{\mathrm{t1}} = x_{\mathrm{b}}$ and $\hat{x}$ should approximate $x_j = x_{\mathrm{t2}}$. At test time, we can render new scenes $\hat{x}$ where only $x_{\mathrm{t1}}$ and $x_{\mathrm{t2}}$ need to be from the same sequence.}
    \label{fig:explain}
\end{figure}

To be more specific, assume an input video of a dynamic scene captured by a static camera. The background consists of the objects that are constant across the frames.
In contrast, the foreground consists of the objects that are equivariant with respect to some family of transformations. We use this difference to infer a scene representation that captures separately the background and the characters.
Building on Dupont et al.~\cite{dupont2020equivariant}, we attain two novel results:

\begin{enumerate}
    \item \textbf{Learn the transformation} We do not require the underlying transformation applied to foreground objects during training, but instead learn it from nearby video frames. At inference, this allows us to animate characters according to any transformation and not just ones that include the character being transformed.
    \item \textbf{Distinguish characters and background}: By independently encoding the background and the manipulable character, we yield disentangled representations for both and can mix and match them freely.
\end{enumerate}

Our model is trained in a self-supervised fashion requiring only nearby frames $(x^{i}_p, x^{i}_q)$ (without any annotation) from the $i$th video sequence. We impose no other constraint and can render new scenes combining characters and backgrounds on the fly with additional (potentially unrelated) frames from sequences $x^{j}$ and $x^{k}$. 
In Sections~\ref{sec:movingmnist} and \ref{sec:sprites} respectively, we show strong results on both Moving MNIST~\cite{DBLP:journals/corr/SrivastavaMS15} with static backgrounds and 2D video game sprites from the Liberated Pixel Cup~\cite{bart_2012} where we demonstrate the following manipulations: 
\begin{itemize}
    \item Render the character in $x^{i}_m$ but with the background from $x^{j}_n$.
    \item Render the character in $x^{i}_m$ using the transformation seen in the change from $x^{j}_p \rightarrow x^{j}_q$.
    \item Combine both manipulations to render the character in $x^{i}_m$ on the background from $x^{j}_n$ using the transformation exhibited by the change seen from $x^{k}_p \rightarrow x^{k}_q$.
\end{itemize}

In Section~\ref{sec:fashion}, we report results on Fashion Modeling, a more realistic dataset for which our results are not as mature.



%% file: core/3_background.tex
\section{Background}
\label{sec:background}

Following Dupont et al.~\cite{dupont2020equivariant}, we define an image $x \in X = \mathbb{R}^{c \times h \times w}$ and a scene representation $z \in Z = \mathbb{R}^{c' \times h' \times w'}$, a rendering function $g: Z \rightarrow X$ mapping scene representations to images, and an inverse renderer $f: X \rightarrow Z$ mapping images to scene representations. It is helpful to consider $x$ as a 2D rendering of a character $c$ from a specific camera viewpoint. An equivariant scene representation $z$ satisfies the following relation for an equivariant renderer $g$ with respect to transformation $T$:
\begin{equation}
\label{eqn:equivariant}
    T^{X}g(z) = g(T^{Z}z)
\end{equation}


\begin{figure}[t] 
    \centering
    \includegraphics[width=.37\textwidth]{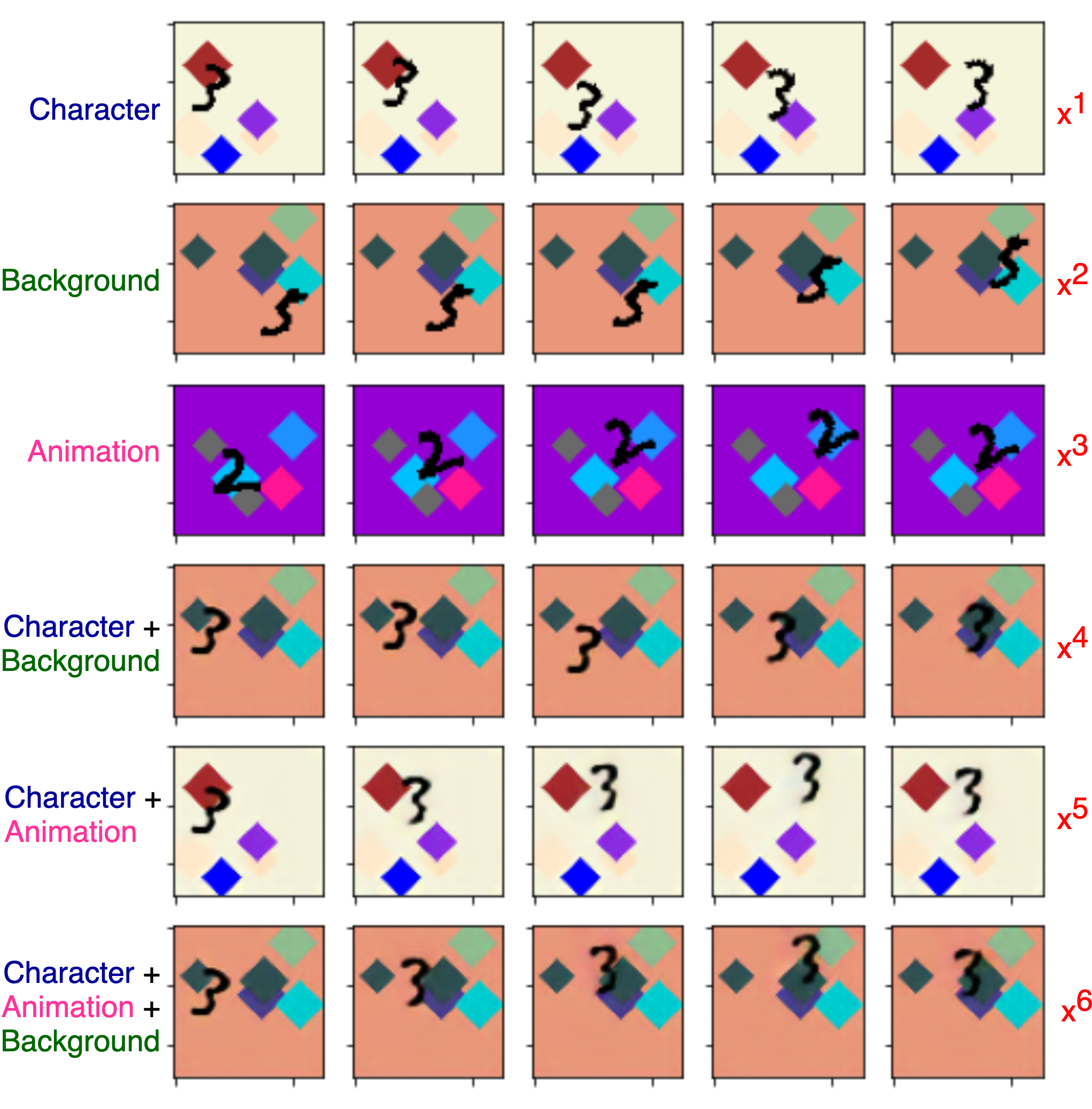}
    \caption{Labeling the rows as $x^1, \ldots, x^6$, the first three are ground truth, row $x^4$ pairs the background from $x^2$ and the character from $x^1$, row $x^5$ pairs the character of $x^1$ with the animations from $x^3$ and row $x^6$ combines both manipulations. With counterclockwise direction and bottom left origin, the $x^3$ transformations are rotate(15), rotate(9), translate(10, 4), translate(-8, -6).}
    \label{fig:allmanipulations}
\end{figure}

In other words, transforming a rendering in image space is equivalent to first transforming the scene in feature space and then rendering the result. Dupont et al. posit functions $f$ and $g$ satisfying the following, where $x$ is a 2D image rendering, $z$ is the scene representation of $x$, and $T^Z$ is an affine transformation:
\begin{equation}
\label{eqn:autoencoder}
    x_2 = T^{X}x_1 = T^{X}g(z_1) = g(T^{Z}z_1) = g(T^{Z}f(x_1))
\end{equation}

Equation~\ref{eqn:autoencoder} is a difficult equality to satisfy, so they propose learning neural networks $f, g$ to approximate it from data triples $(x_1, x_2, \theta)$, where $x_1, x_2$ are renderings of the same character $c$ and $\theta$ is a rotation transforming $c$ from its appearance in $x_1$ to its appearance in $x_2$. They assume $T^Z$ is the same operation as $T^X$ but in feature space, and consequently apply a rotation of $\theta$ to the 3D representation $z_1 = f(x_1)$. With $\tilde{z}_1 = R^{Z}_{\theta}f(x_1)$ and $\tilde{z}_2 = (R^{Z}_{\theta})^{-1}f(x_2)$, they then train $g$ and $f$ to minimize reconstruction loss:
\begin{align}
    \mathbb{L}_{\text{render}} = ||x_2 - g(\tilde{z_1})|| + ||x_1 - g(\tilde{z_2})||
    \label{eqn:loss}
\end{align}

After training, new renderings can be inferred by first inverse rendering $z = f(x)$, then applying rotations $T^Z$ upon $z$, and finally rendering images $\hat{x} = g(T^{Z}z)$. In other words, the model operates entirely in feature space. For their purposes, this lets them manipulate the output rendering quickly and without access to the object in image space. As we show in Section~\ref{sec:method}, this can be further extended to manipulations that are actually very difficult to do in image space but become easy to do in feature space.

%% file: core/4_method.tex
\section{Method}
\label{sec:method}

Our motivation is in using the smoothly-changing nature of video to learn the transformations between frames. Like Dupont et al., we assume that the change between frames $F_t$ and $F_{t+1}$ can be modeled with affine transformations. While they use only rotation, we assume an arbitrary affine transformation on the character plus an invariant transformation on the background, in-painting as needed. Accordingly, we remove the $\theta$ requirement at both training and test by learning the transformation $T^Z$ from data. A question arises -- if we learn $T^Z$ from data, then why assume that it is an affine operation? 
The first reason is that it biases $T^Z$ towards having a commonly used model for motion estimation.
Second, it limits us to six interpretable axes of variation in our animation, which lets us compare directly against a ground truth affine transformation when given. 
Finally, this approach also limits the operator's ability to `hide' information about the target image in the representation, exemplified by works like CycleGAN~\cite{CycleGAN2017} as demonstrated by Chu et al.~\cite{DBLP:journals/corr/abs-1712-02950}.

\paragraph{Setup} Building on 
Section~\ref{sec:background}, we define $f_b$ and $f_c$ respectively representing the encoding of the background and character. With scalar coefficients $\alpha_{\text{equiv}}$, $\alpha_{\text{inv}}$, and denoting
\begin{align*}
    & h(x_{\mathrm{c}}, x_{\mathrm{t1}}, x_{\mathrm{t2}}, x_{\mathrm{b}})\\
    \vspace*{2pt}
    & \quad= g(T^{Z}(f_{c}(x_{\mathrm{t1}}), f_{c}(x_{\mathrm{t2}})) \circ f_{c}(x_{\mathrm{c}}) + f_{b}(x_{\mathrm{b}}))
\end{align*}
we learn neural networks $f_c$, $f_b$, $g$, and $T^Z$ by minimizing $\mathbb{L}_{\text{total}}$ below. During training, this requires a dataset consisting of random pairs $(x_1, x_2)$ of frames from each clip where $x_1$ and $x_2$ have roughly the same background.
\begin{align}
        \mathbb{L}_{\text{total}} &= \mathbb{L}_{\text{scene}} + \alpha_{\text{equiv}}\mathbb{L}_{\text{equiv}} +
    \alpha_{\text{inv}}\mathbb{L}_{\text{inv}} 
\end{align}
where
\begin{align*}
    \mathbb{L}_{\text{scene}} &= ||h(x_1, x_1, x_2, x_1) - x_2||_2 \\
    \mathbb{L}_{\text{equiv}} &= ||T^{Z}(f_{c}(x_1), f_{c}(x_2)) \circ f_{c}(x_1) - f_{c}(x_2)||_2 \\
    \mathbb{L}_{\text{inv}} &= ||f_b(x_1) - f_b(x_2)||_2
    \end{align*}



Losses $\mathbb{L}_{\text{equiv}}$ and $\mathbb{L}_{\text{invariant}}$ help with training, but they are not by themselves indicative of successful training. We attain quality results only when $\mathbb{L}_{\text{scene}}$ is sufficiently minimized. Other possible constraints, such as a transformation inverse loss on $T^{Z}(f_{c}(x_{\mathrm{t1}}), f_{c}(x_{\mathrm{t2}})) = T^{Z}(f_{c}(x_{\mathrm{t2}}), f_{c}(x_{\mathrm{t1}}))^{-1}$, were not necessary.

Both $T^Z$ and $f_c$ learn to handle every character similarly. This is important because it means that at inference time we can render novel scenes given a pair of nearby frames in a video. The renderings described in Section~\ref{sec:intro} and demonstrated in Section~\ref{sec:experiments} are now:

\begin{itemize}
    \item $h(x^{i}_1, x^{i}_1, x^{i}_2, x^{j}_1)$: Render the character in $x^{i}_1$ as seen in $x^{i}_2$ but with the background from $x^{j}_1$.
    \item $h(x^{j}_1, x^{i}_1, x^{i}_2, x^{j}_1)$: Render the character and background in $x^{j}_1$ using the transformation exhibited by the change in the character from $x^{i}_1 \rightarrow x^{i}_2$.
    \item $h(x^{i}_1, x^{k}_1, x^{k}_2, x^{j}_1)$:
    Combine the above two to render the character in $x^{i}_1$ on the background from $x^{j}_1$ using the transformation exhibited by the change in the character from $x^{k}_1 \rightarrow x^{k}_2$.
\end{itemize}

\paragraph{Learning the transformation.} The smoothly changing nature of video lets us advance past affine transformations parametrically defined with a $\theta$ angle of rotation and instead learn transformations based on frame changes. One advantage to this approach is that the model becomes agnostic to which transformations the data exhibits. We show this to be explicitly true for Moving MNIST and implicitly true for Sprites, where the transformations are not affine. Another advantage arises at inference time because the model is agnostic to whether the frame to be transformed is input to the transformation function. 

\paragraph{Distinguishing characters and background} The renderer $g$ is equivariant to character transformations. A static background however will be constant across a video. We take advantage of this to learn $g$ along with functions $f_c$ and $f_b$ corresponding to the character and the background such that, at inference time, we can mix and match characters and backgrounds previously never seen together.

%% file: core/5_experiments.tex
\section{Experiments}
\label{sec:experiments}

The first dataset is a variant of Moving MNIST~\cite{DBLP:journals/corr/SrivastavaMS15} where we overlay each sequence of moving digits on a randomly drawn background. Experiments on this dataset lets us test our model on a wide range of explicit translations and rotations in the digits. 

The second dataset was debuted by Reed et al.\cite{NIPS2015_5845} and consists of moving 2D video game character sprites without any backgrounds, using graphic assets from the Liberated Pixel Cup~\cite{bart_2012}. We perform experiments both with and without backgrounds. These experiments test our model on more complex mappings like firing a bow and arrow instead of just affine transformations in image space.


This test-bed is suitable for demonstrating that our method exhibits three desirable capabilities: 1) Separate foreground from background; 2) Learn an interpretable but complex transformation of the foreground in the feature space; 3) Render transformed foreground objects. Furthermore, it allows us to demonstrate generalization to unseen characters (digits and sprites) and unseen backgrounds.

We also show results on a third dataset, Fashion Modeling, from Zablotskaia et al.~\cite{DBLP:journals/corr/abs-1910-09139}, consisting of videos of models exhibiting clothing without any backgrounds. This dataset tests our model on realistic poses and fine detail.

\subsection{Architecture} 

We parameterize $f_c$, $f_b$, $g$, and $T^Z$ with neural networks. While separate, both $f_c$ and $f_b$ share the same architecture details as a residual network. The renderer $g$ is a transposition of that architecture, although we drop the residual components. Please see Figures~\ref{fig:encoder_architecture},~\ref{fig:decoder_architecture},~\ref{fig:siamese_architecture} in the Appendix for details.

The transformation $T^Z$ exhibits three properties. First, it is input-order dependent. Second, it uses PyTorch's~\cite{paszke2017automatic} \texttt{affine\_grid} and \texttt{grid\_sample} functions to transform the scene similarly to how Spatial Transformers~\cite{DBLP:journals/corr/JaderbergSZK15} operate. The third is that it is initialized as identity, reflecting our prior that $T^Z$ does not alter the foreground.

\subsection{Background Generation}
\label{sec:backgrounds}

We create $B$ randomly generated backgrounds for each of train, val, and test. For each background, we select a color from the Matplotlib CSS4 colors list. We then place five diamonds on the background, each with a different random color, along with an independent and randomly chosen center and radius. The radius is uniformly chosen from between seven and ten, inclusive. 

For the Moving MNIST experiments, $B=64$ for each split because the model overfit with $B < 64$. For the Sprite experiments, we used $B = 100$ for test but a range of $B \in [64, 128, 256, 784, 1000]$ for training.


\subsection{Moving MNIST} 
\label{sec:movingmnist}

We generate videos, each of length $M=5$, of MNIST digits (characters) moving on a static background. The digits and background have dimensions $(28, 28)$ and $(64, 64)$ respectively. At each training step, we select $N$ digits in the train split of MNIST, as well as a background from the set of pre-generated training backgrounds. We then randomly place these digits. For each digit, and for each of $M-1$ times, we choose randomly between rotation and translation. If translation, then we translate the character independently in each of the $x$ and $y$ directions by a random amount in $\{-10, -8, \ldots, 8, 10\} \setminus \{0\}$. If rotation, then we rotate the character by a random amount in $\{-15, -12, \ldots, 12, 15\} \setminus \{0\}$. If the character leaves the image boundaries, we redo the transformation selection. Otherwise, that transformation is applied cumulatively to yield the next character position. 

This results in digits on blank canvases. We overlay them on the chosen background to produce a sequence where the change in each character from frames $F_t \rightarrow F_{t+1}$ is small and affine for the character and constant for the background. This is performed by locating the (black) MNIST pixels and blackening those locations on the canvases. Accordingly, we do not use the black color canvas. We then randomly choose two indices $i, j$ and use $(x_i, x_j)$ as the training pair. See Figure~\ref{fig:allmanipulations} for example sequences.



\paragraph{Qualitative Results} Our model learns to render new scenes using characters from the test set of MNIST and held out backgrounds. All shown sequences are on unseen backgrounds with unseen MNIST digits where there are at least two transformations of each of rotation and translation.

Figure~\ref{fig:allmanipulations} convincingly demonstrates the manipulations from Section~\ref{sec:method}. The final row is rendered by encoding the character from the first row, the background from the second row, and using the transformations exhibited in the third row. Denoting $x^{i}_{j}$ as the $j$th frame from the $i$th sequence: $$x^{6}_{k} = h(x_{\mathrm{c}} = x^1_k, x_{\mathrm{t1}} = x^3_1, x_{\mathrm{t2}} = x^3_k, x_{\mathrm{b}} = x^2_k)$$

\begin{figure}[t] 
    \centering
    \includegraphics[width=.35\textwidth]{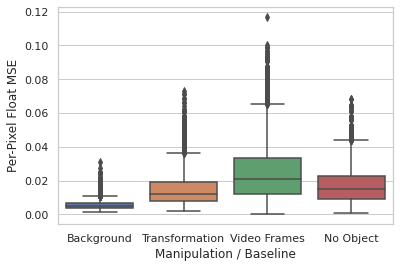}
\caption{Per-pixel MSE over 10,000 test examples. The transform and background manipulations use our learned functions; Video frames is MSE of a frame against a random (non-identical) frame from the same video; No object is MSE of the background versus the full frame of background and character.}
    \label{fig:pixelmse}
\end{figure}

\paragraph{Quantitative Results} We tested reconstruction by evaluating the per pixel MSE over the moving MNIST test set. For each example, we randomly chose two pairs of (background, digit) and made corresponding videos $(x^1_1, \ldots, x^1_5)$ and $(x^2_1, \ldots, x^2_5)$. We then indexed into the same random position in both sequences to get frame pairs $(x^1_i, x^1_j)$, $(x^2_i, x^2_j)$. Fig~\ref{fig:pixelmse} shows a boxplot comparing two manipulations, \textit{Transformation} and \textit{Background}, along with two baselines, \textit{Video frames} and \textit{No object}. 

The predicted value for \textit{Transformation} is $\hat{x}^2_{j} = h(x^2_i, x^1_i, x^1_j, x^2_i)$. We compute an MSE for this manipulation by comparing it against the ground truth attained by rendering the background and character from $x^2_{i}$, transformed like seen in $x^1_{i} \rightarrow x^1_{j}$. The predicted value for \textit{Background} is $\hat{x}^1_{j} = h(x^1_i, x^1_i, x^1_j, x^2_j)$. We compute an MSE for this manipulation by comparing it against the rendering of the character in $x^1_{j}$ on the background from $x^2_{j}$ with the original transforms. \textit{Video frames} is the MSE of two random frames from the same video. \textit{No object} is the MSE of a full frame against only the background from that frame. 

Given that MSE is a measure of reconstruction quality with lower values being better, we expect them to serve as upper bounds. \textit{Video frames} is the upper bound when reconstruction gets the character but places it incorrectly. \textit{No object} is the upper bound when reconstruction fails to include the character. On this measure, we see that the background manipulation is much better than the baselines, but we cannot say with certitude that the transform manipulation is better as it is within confidence interval of \textit{Video frames} and its boxplot overlaps with both baselines.

\begin{figure*}[!ht]
    \centering
    \subfloat[\centering Spellcast Animation]{{\includegraphics[width=5.25cm]{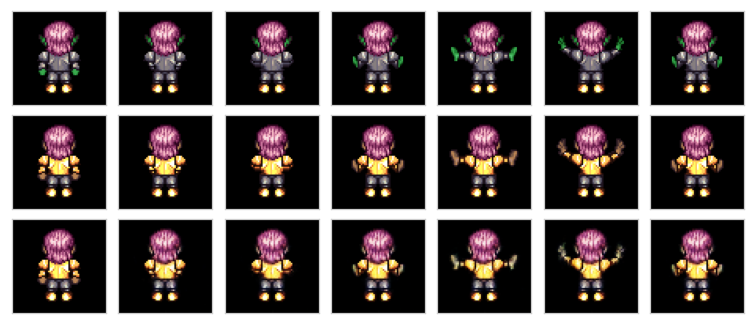} }}%
    \qquad
    \centering
    \subfloat[\centering Thrust Animation]{{\includegraphics[width=5.25cm]{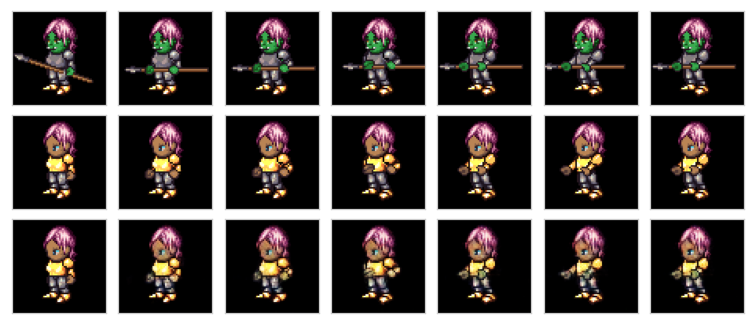}}}
    \qquad
    \centering
    \subfloat[\centering Shoot Animation]{{\includegraphics[width=5.25cm]{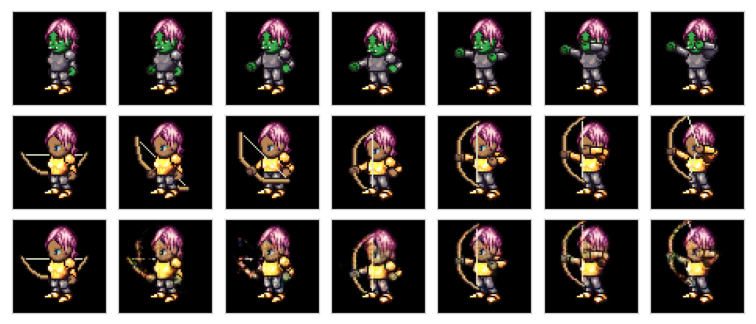}}}
    \caption{Transferring animations. In both panels, the first and second rows are ground truth; Image $x^3_i$ in the third row is given by applying the transformation exhibited by $x^1_1 \rightarrow x^1_i$ to $x^2_0$. Observe how in panel (b) that a spear is not hallucinated, but instead the thrust action is applied to the character without a spear. And in panel (c), even though there is no bow in the input to the transformation, the model understands to move the bow and pull the string in the character upon which the transformation is applied.}%
    \label{fig:sprite-animations}%
\end{figure*}

\begin{figure}[!h]
    \centering
    \includegraphics[width=0.5\textwidth]{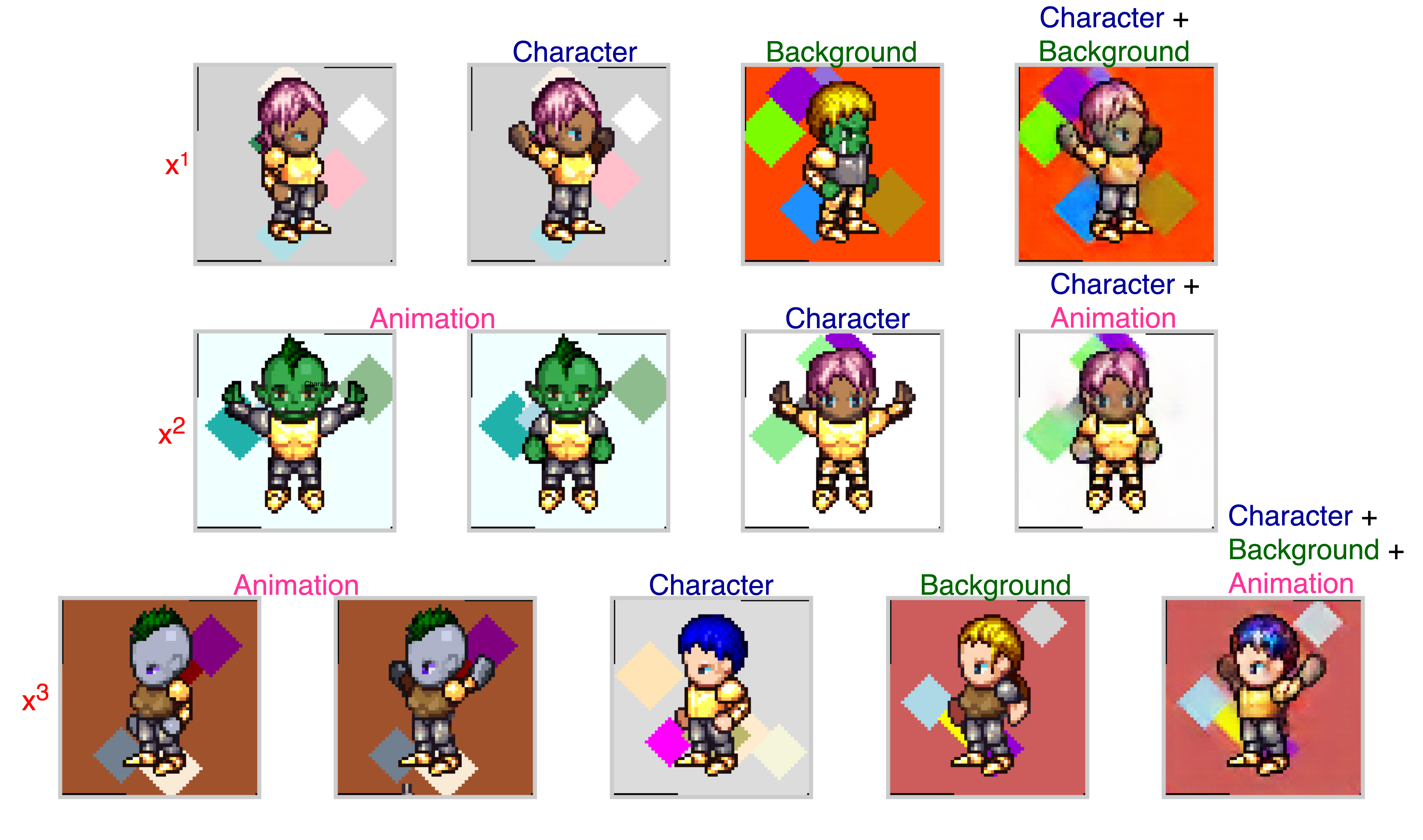}
    \caption{The rows are background manipulation $x^1_4 = h(x^1_1, x^1_1, x^1_2, x^1_3)$, animation transfer $x^2_4 = h(x^2_3, x^2_1, x^2_2, x^2_3)$, and both combined $x^3_5 = h(x^3_3, x^3_1, x^3_2, x^3_4)$.}
    \label{fig:sprite-animations-background}%
\end{figure}


\subsection{Video Game Sprites} 
\label{sec:sprites}

The dataset comprises $60 \times 60$ color images of sprites with seven attributes (sex, body type, arm, hair, armor, greaves, and weapon) for a total of 672 unique characters. We use the same dataset splits as Reed et al. with 500 training characters, 72 validation characters, and 100 test characters. For each character, there are four viewpoints for each of five animations: spellcast, thrust, walk, slash, and shoot. Our sequences are the twenty animations per character, which have between six and thirteen frames. We only show results on test characters and a held out set of test backgrounds.

For results without backgrounds, we select a random sequence and then two random frames $(x_i, x_j)$ from that sequence as input to our model. For results with backgrounds, we additionally choose a random background and then use the provided masks to center the sprite data on top of the background. See Figures~\ref{fig:sprite-animations}, \ref{fig:sprite-animations-background} for examples of sequences without and with backgrounds.

\paragraph{Qualitative Results} Figure~\ref{fig:sprite-animations} shows strong results on three different types of animations (spellcast, thrust, shoot) without backgrounds. In each panel, the first and second rows are ground truth, as is the first frame of the third row -- $x^3_1$ ($= x^2_1$). The rest of the third row is animating $x^3_1$ according to the transformations exhibited by the first row. In all three panels, the major concern is with the hand color. The model occasionally blends the green hands of the original character with that of the character being animated. In the first panel, this appears as gray, while in the second and third panels it has a green tint. On the other hand, the model successfully performs the thrusting action without hallucinating a spear in the second panel. Similarly, the model operates the bow and arrow in the third panel, even though there is only the shooting motion and not the actual bow in the first row.

Figure~\ref{fig:sprite-animations-background} shows results when using backgrounds. We see an example of each of the type of manipulations from Section~\ref{sec:method}. The first row is changing the background, the second row is animation transfer, and the third row is performing both background change and animation transfer. The model is strong but clearly not perfect at this task. In addition to the hand discoloring (2nd row), we also see that the backgrounds are a bit blurry near the character in all three rows.

\begin{figure}[!ht]
    \centering
    \includegraphics[width=6.25cm]{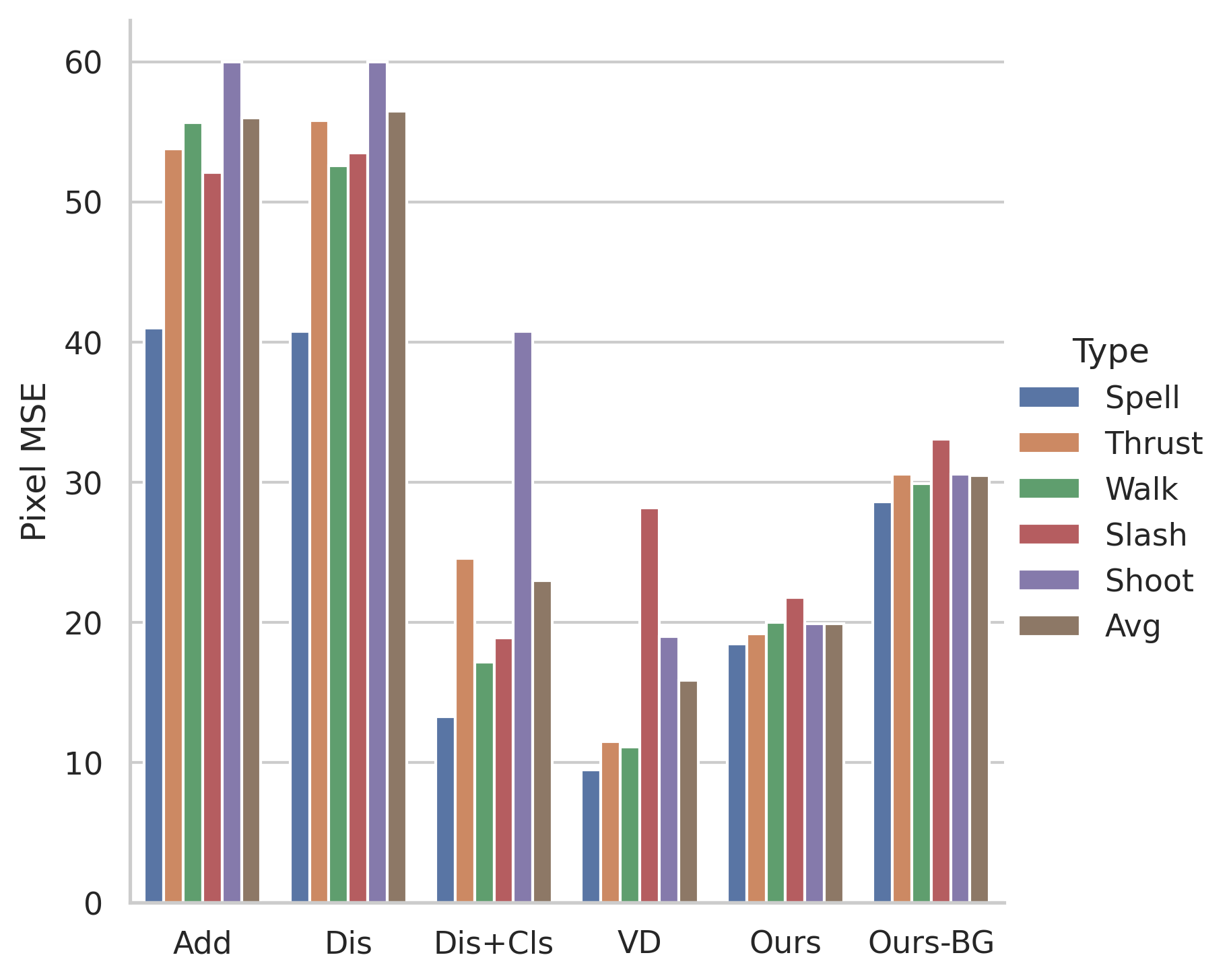}%
    \caption{Reconstructing analogous sprites without backgrounds in pixel MSE, with max cutoff at 60.0. Our method is competitive in every category and has much lower variance than the other models, even though it was designed for datasets with affine transformations and backgrounds, both of which are missing in this context. We highlight our model's flexibility by using a model trained with diamond backgrounds (`Ours-BG'), asking it to remove the background, and still yielding comparable results.}%
    \label{fig:analogyrecon}%
\end{figure}

\begin{figure}[!h]
    \centering
    \includegraphics[width=5.25cm]{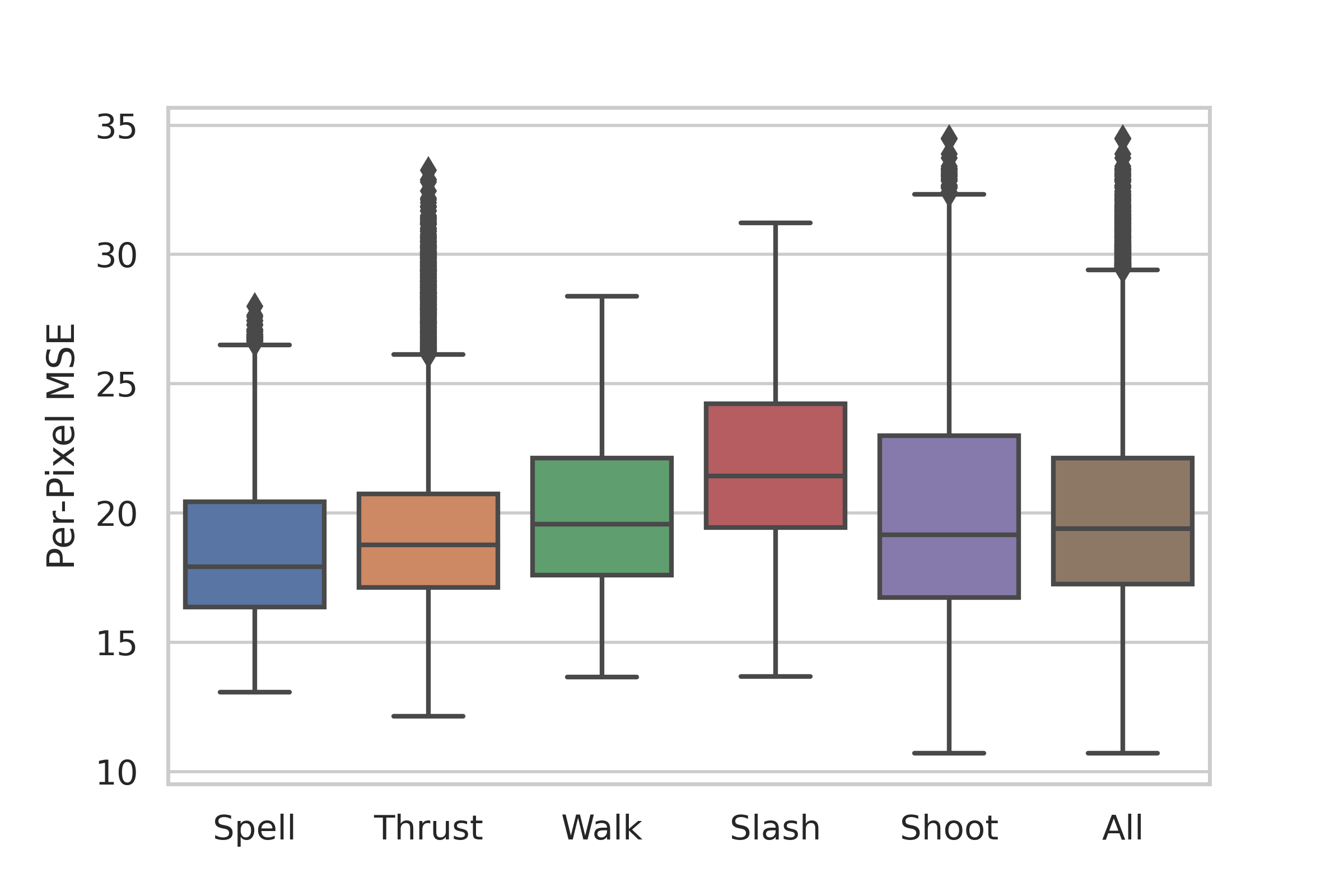}%
    \caption{A boxplot of sprite analogy reconstruction MSE without backgrounds. This expands upon the reported averages in Figure~\ref{fig:analogyrecon}.}%
    \label{fig:sprite-bw-recon}%
\end{figure}

\paragraph{Quantitative Results} For results without backgrounds, we tested reconstruction by evaluating the per pixel MSE over the test set. Figure~\ref{fig:analogyrecon} compares our approach against prior work on analogy reconstruction. We are competitive in every category and the most consistent in our results, with Figure~\ref{fig:sprite-bw-recon} showing that our model is capable across all categories. This is in contrast to the model with the best average performance from Xue et al.~\cite{DBLP:journals/corr/XueWBF16}, `VD', which performs poorly on the slash category. We further show our model's flexibility by taking a version trained with diamond backgrounds and ask it to perform the same analogy task but with a fixed gray canvas as the input to the background encoder. We use gray because the model never saw black in its training process. Figure~\ref{fig:analogyrecon} shows that this model, `Ours-BG', attains comparable results. Examples from `Ours-BG' and a representative MSE boxplot are in the Appendix (Figures~\ref{fig:recon-removeback-mse},~\ref{fig:recon-removeback-examples}).

For sprite results with backgrounds, we similarly evaluate per-pixel MSE evaluation in Figure~\ref{fig:sprite-color-recon}. We include only the aggregate boxplots and put the full breakdown in the Appendix (Figure~\ref{fig:sprite-color-recon-breakdown}). The first boxplot, `Analogy', represents analogy reconstruction. It shows a clear (expected) increase in MSE over reconstruction without backgrounds. Is this due to the background or the character? 

To answer that question, we utilize the provided character mask. Boxplot `Background-only' is the resulting MSE from masking out where the character should be in both the predicted and ground truth and then computing MSE over just the unmasked part, presumed to be the background. Boxplot `Character-only' is similar but masking out everything but where the character should be and computing MSE over the presumed character. `Background-only' MSE remains consistent but is also much lower than `Analogy', with an aggregate of $28.5$. In contrast, the `Character-only` aggregate MSE is $79.0$ with a high of $83.3$ (slash) and a low of $73.3$ (spell). This analysis suggests that the loss in MSE is dominantly due to the character reconstruction. This is encouraging for improving our results given the litany of recent techniques~\cite{siarohin2020order,siarohin2020motionsupervised,DBLP:journals/corr/abs-1910-09139,Tulyakov_2018_CVPR} focusing on character animation in similar settings that can be coupled with our method.


\begin{figure}[!h]
    \centering
    \includegraphics[width=6.25cm]{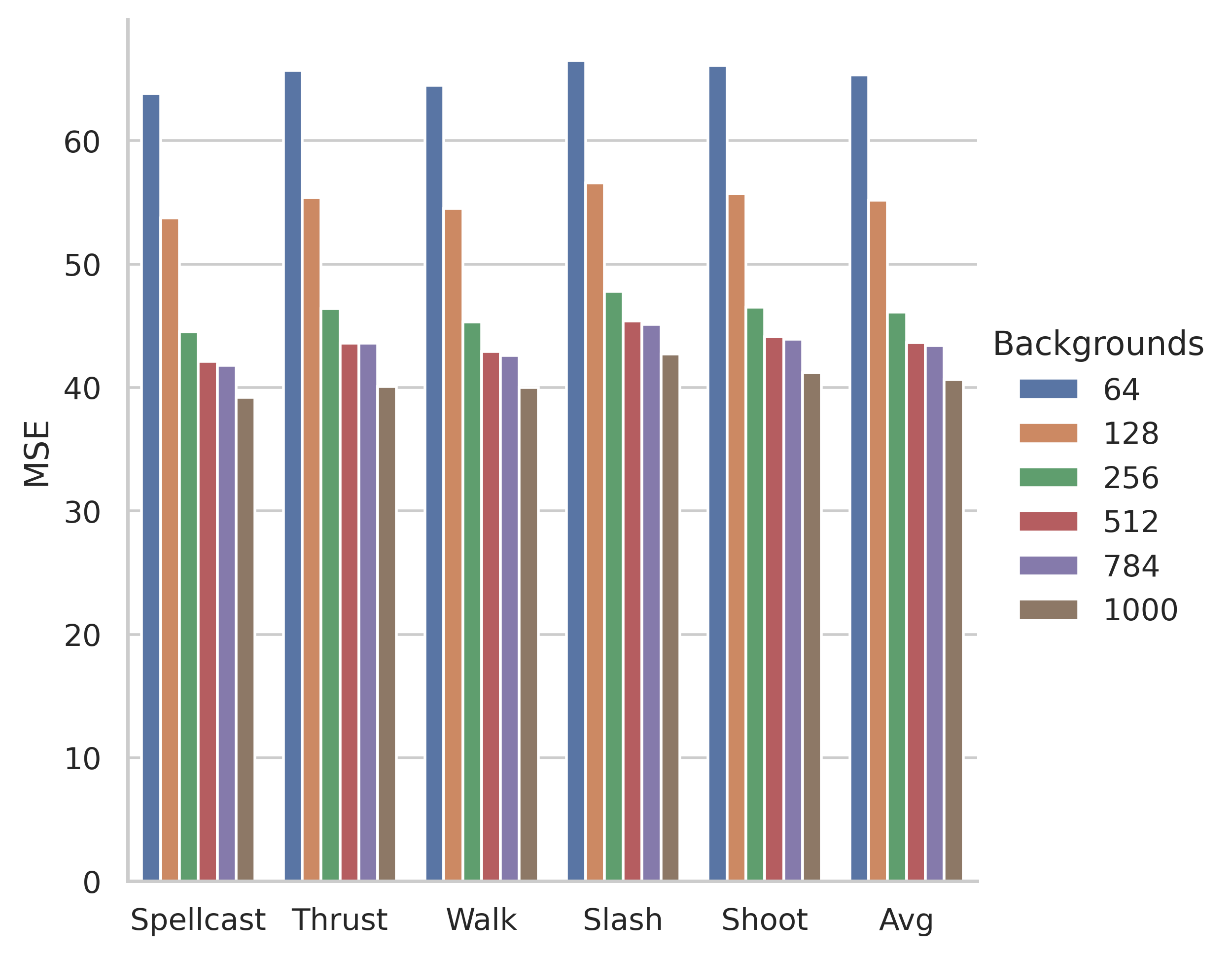}%
    \caption{Analogy reconstruction with backgrounds. The more the number of backgrounds used during training, the better the model does at test set reconstruction. This effect tapers near 1000.}%
    \label{fig:sprite-results-background}%
\end{figure}

\begin{figure}[!h]
    \centering
    \includegraphics[width=6.25cm]{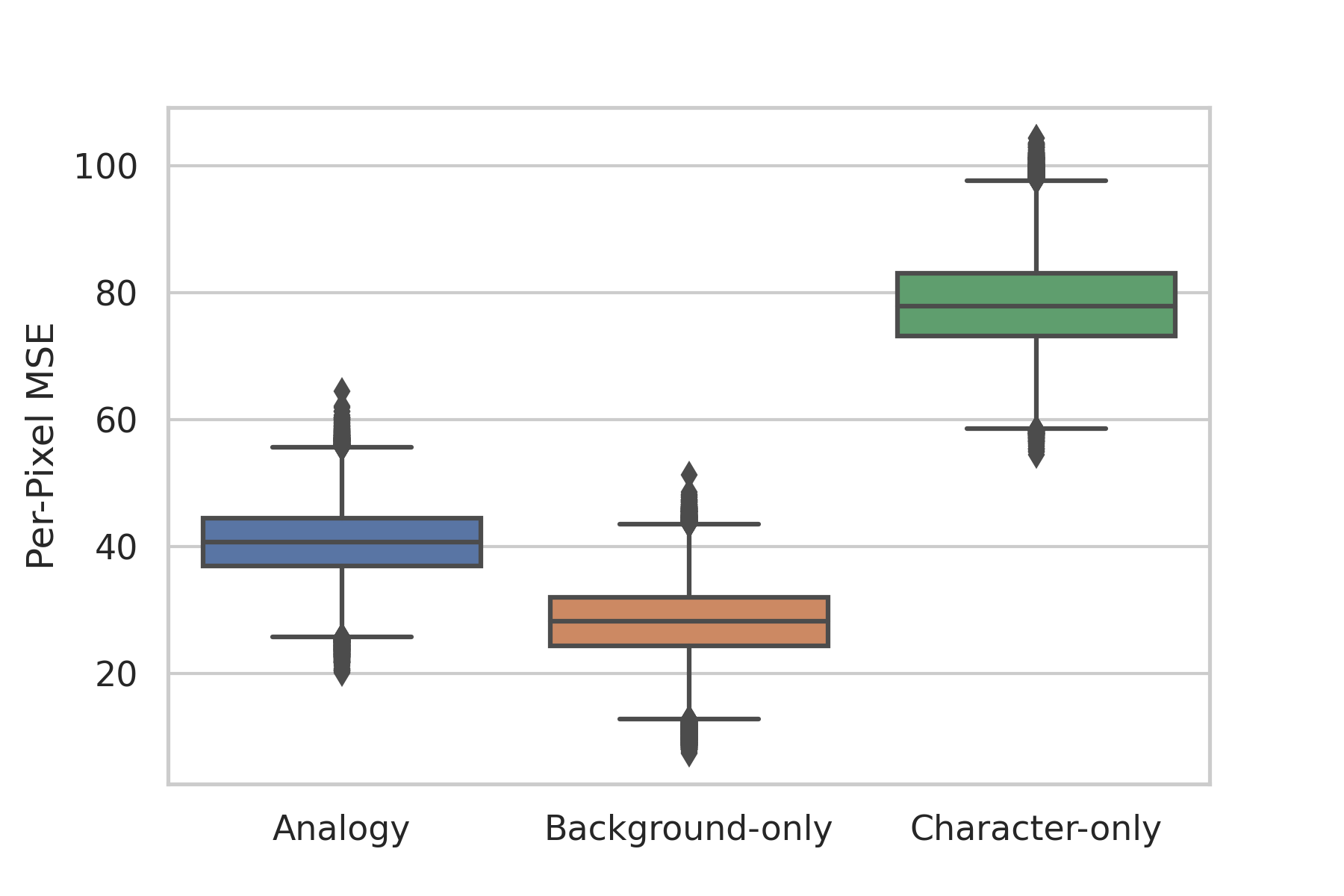}
    \caption{Sprite analogy reconstruction with backgrounds over all animation types. Compared to Figure~\ref{fig:sprite-bw-recon}, the `Analogy' boxplot shows an (expected) increase in MSE. The other two boxplots highlight that this is dominantly due to character reconstruction rather than background reconstruction.}
    \label{fig:sprite-color-recon}%
\end{figure}

\begin{figure*}[!t]
    \centering
    \includegraphics[width=.41\textwidth]{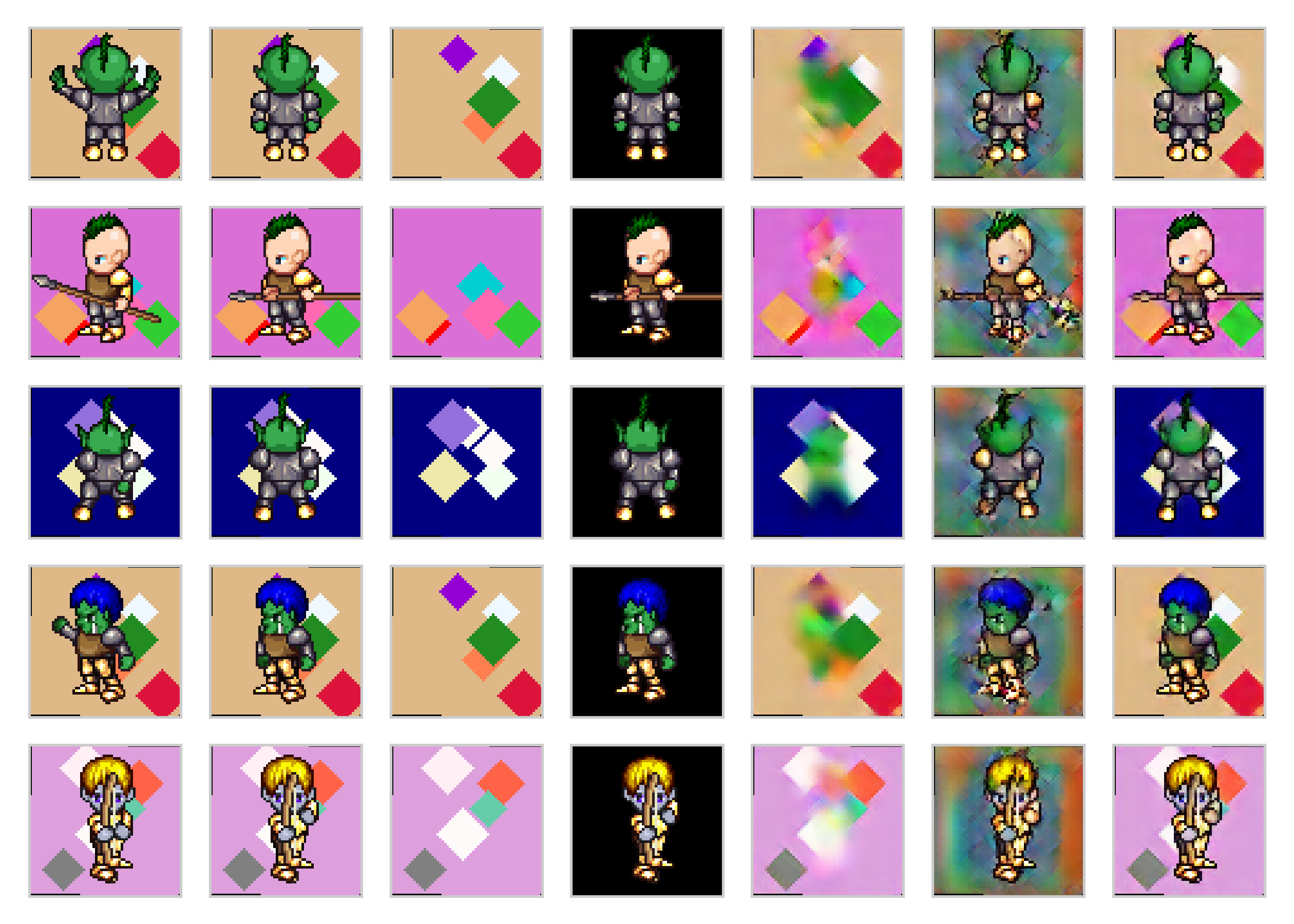}
    \qquad
    \includegraphics[width=.50\textwidth]{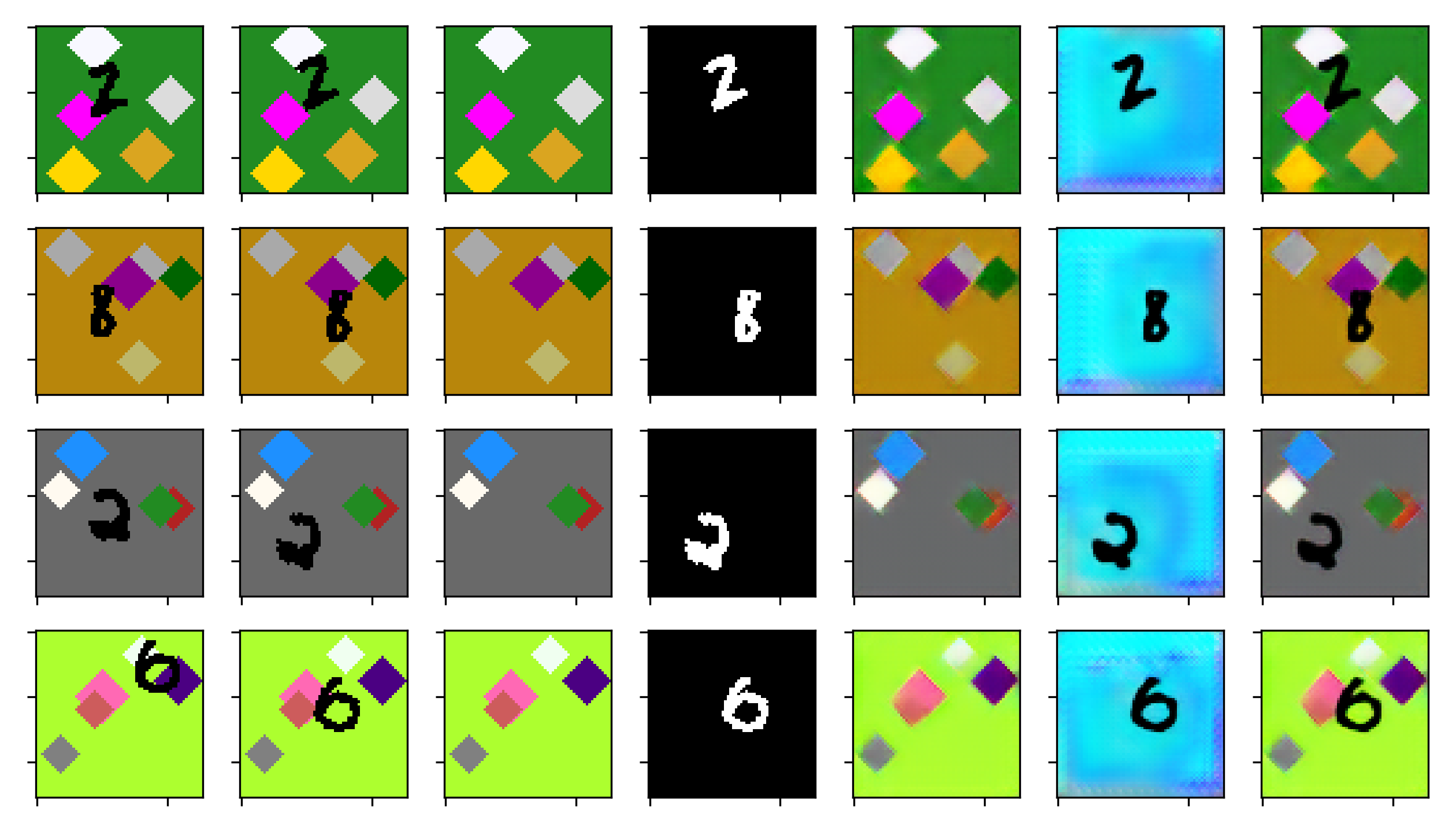}
    \caption{In both panels, the first four columns are ground truth. The fifth is the model's output when inputting just the background encoding to the decoder $g$. The sixth is the output when inputting just the transformed character encoding to $g$. The last column is including both as input to $g$. Notice how the model separates the character and the background more cleanly in Moving MNIST. This is because the characters move throughout the scene and so the model has distinguishing information about all areas of the frame. Figure~\ref{fig:reconboth-212} in the Appendix shows that when the sprite moves during training in only a single cardinal direction, the model learns a similarly clean separation.}
    \label{fig:zerooutboth}%
\end{figure*}

\subsection{Fashion Modeling} 
\label{sec:fashion}

\begin{table}[!h]
\centering
\begin{tabular}{c|ccc}
& Perceptual ($\downarrow$)& FID ($\downarrow$) & AKD ($\downarrow$) \\
\toprule
MonkeyNet~\cite{DBLP:journals/corr/abs-1812-08861} & 0.3726  & 19.74 & 2.47 \\
CBTI~\cite{DBLP:journals/corr/abs-1811-11459} & 0.6434 & 66.50 & 4.20 \\
DwNet~\cite{DBLP:journals/corr/abs-1910-09139} & \bf 0.2811 & \bf 13.09 &  \bf 1.36  \\
Ours & 0.4222 & 70.87 &  13.06 \\
\end{tabular}
\caption{Comparison with state-of-the-art on Fashion.}
\label{table:fashion-sota}
\end{table}

Table~\ref{table:fashion-sota} compares our results on Fashion Modeling against the current state of the art. While our model is capable of learning the transformation, it is not up to par with respect to realism. We consider this to be a question of model architecture and training and are actively working to improve this result. One direction is to adopt the 3D architecture from Dupont et al, which we expect to improve our results greatly given the common output errors. See Figures~\ref{fig:fashion-recon},~\ref{fig:fashion-analogy} in the Appendix for example renderings.

\subsection{Analyzing $T^{Z}$}
\label{sec:analyzing_tz}

There must be structure in $T^Z$ because of its success animating characters. We analyze this with Moving MNIST because we can compare $T^Z$ to the ground truth character transformation in $T^X$. For Sprites, that comparison is more difficult because the ground truth is categorical. We gather $40,000$ unique frame pairs of a fixed digit instance on a fixed background, and show results for 8 and 5. 

\begin{figure}[h]
    \centering
    \includegraphics[width=0.41\textwidth]{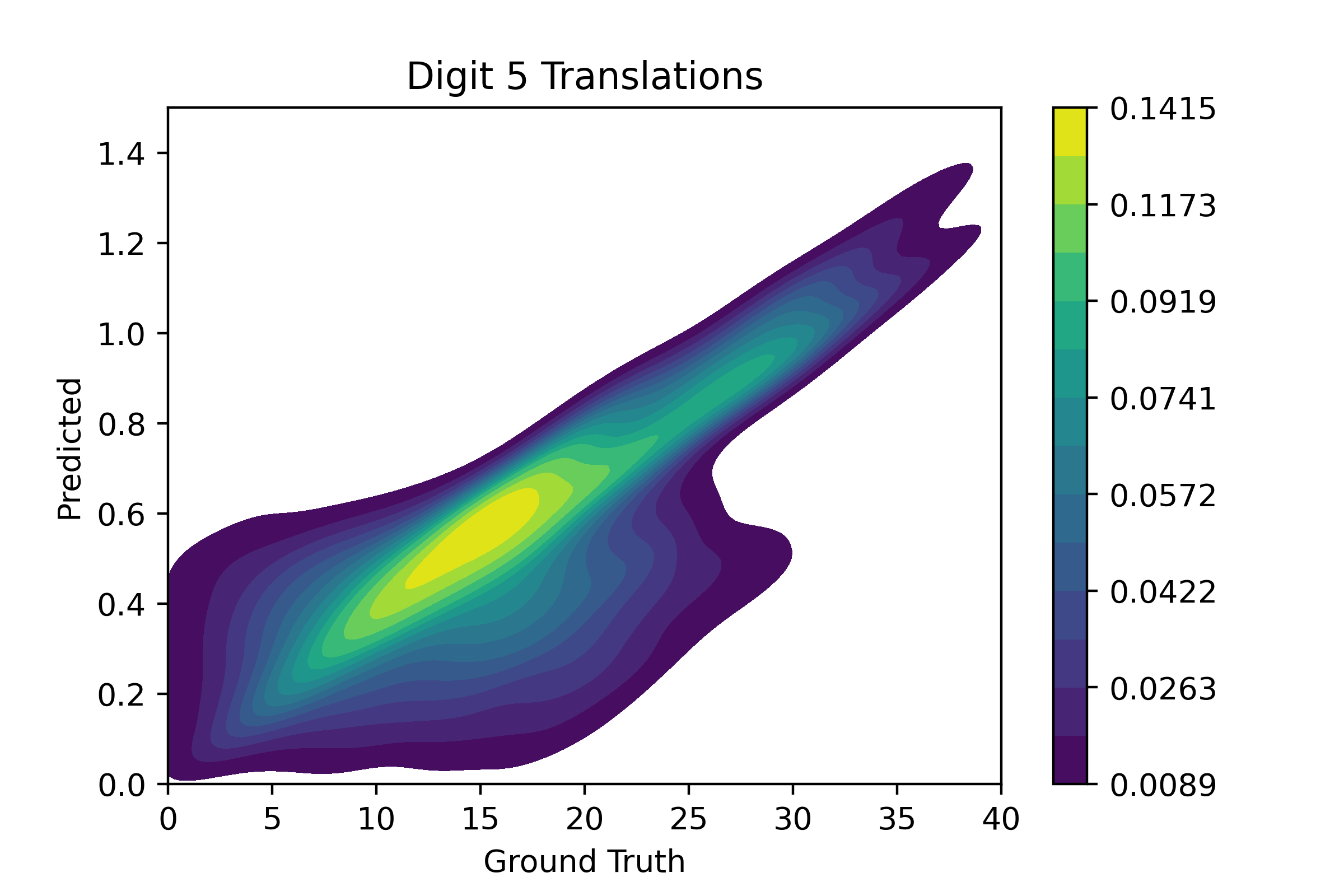}
    \caption{Kernel density estimation plot of the euclidean norm of the translation parameters for Moving MNIST, with $T^Z$ on the y-axis and $T^X$ on the x-axis.}
    \label{fig:tz-translations}%
\end{figure}

We make an assumption that there is structure in the MNIST $T^Z$ with respect to both translation and rotation parameters independently. Figure~\ref{fig:tz-translations} shows a kernel density estimation plot for the euclidean norm of the translation parameters for digit $5$ (see Figure~\ref{fig:tz-translations8} in the Appendix for digit $8$), attained by isolating the third column of the affine matrices. While these are the only parameters corresponding to translation in the gathered $T^X$, we caution that there may be more parameters effecting translation in $T^Z$ due to how affine matrices combine. Observe though how strongly coupled and linear is the relationship between these norms, with almost all of the joint density lying in a small range centered at about $(15, 0.6)$.

\begin{figure}[h]
    \centering
    \includegraphics[width=0.5\textwidth]{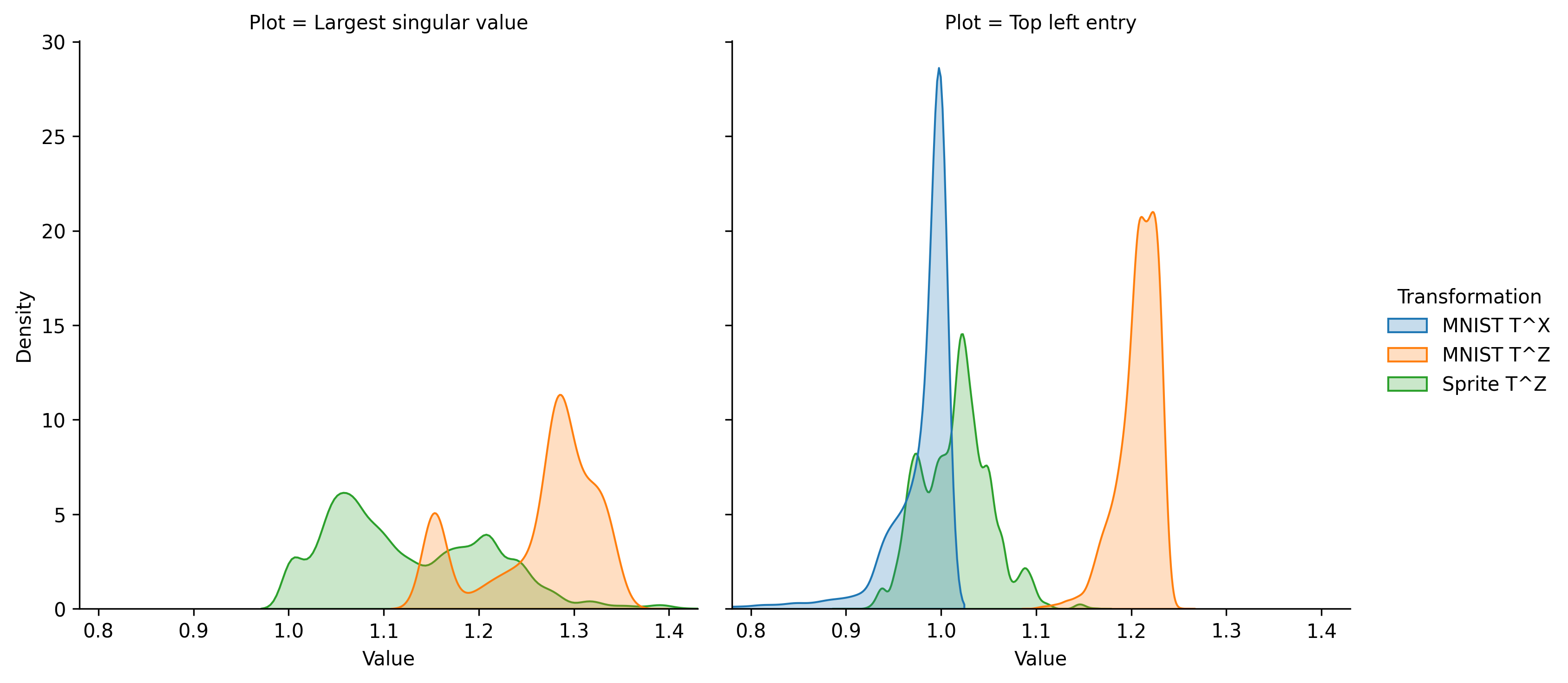}
    \caption{Transformation densities over 40,000 (MNIST) and 105,000 (Sprites) random, unique frame pairs. The left plot is of the largest singular value of the affine transformation, where $T^X$ is not shown as its upper left square matrix is always unitary. The right plot is of the top left entry in the matrix; positive density for $T^X$ value $> 1$ is a plotting artifact.}
    \label{fig:transform-kdeplot}%
\end{figure}

\begin{figure}[h]
    \centering
    \includegraphics[width=0.41\textwidth]{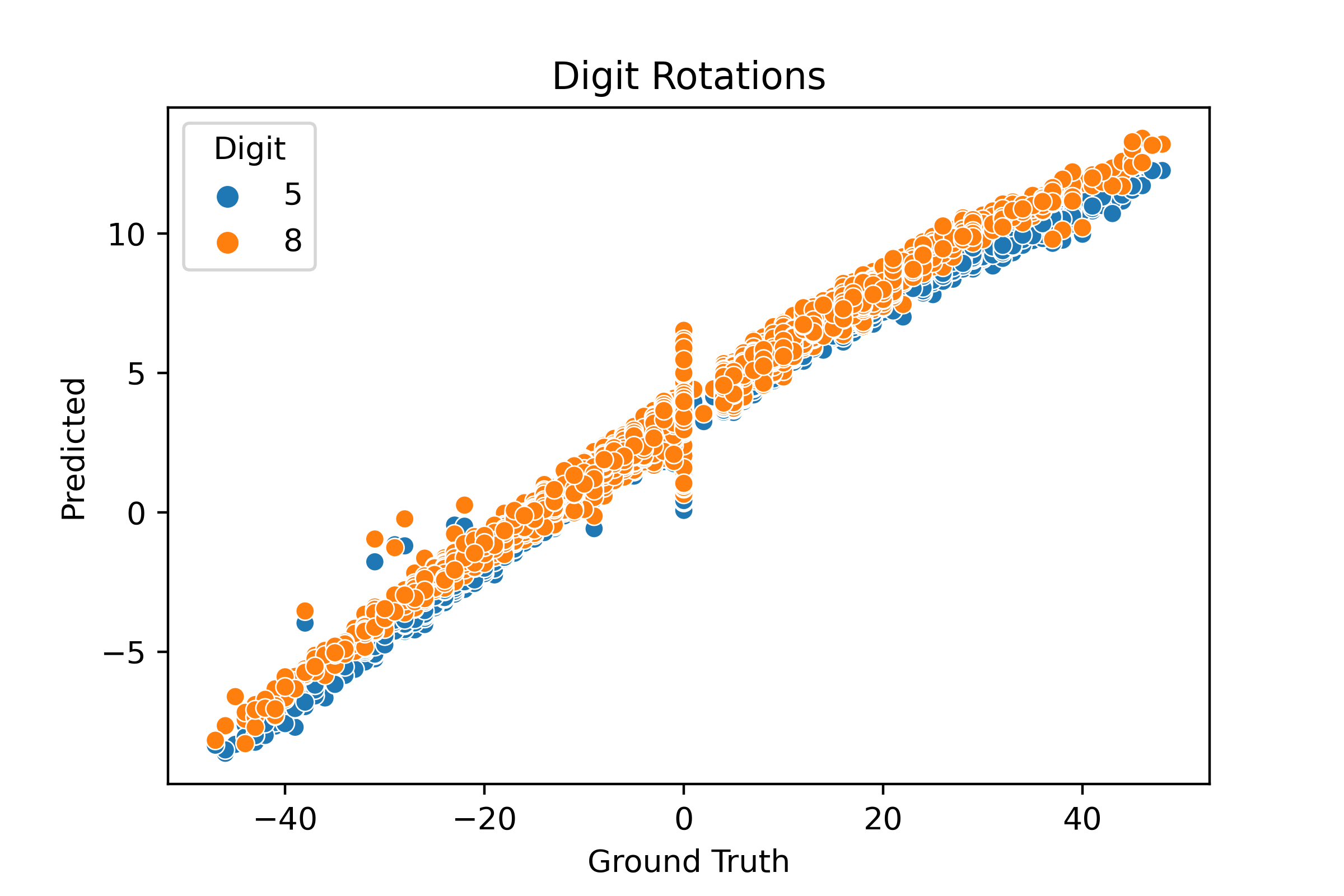}
    \caption{Moving MNIST scatterplot comparing the rotation angle of $T^Z$ against the rotation angle of $T^X$. The latter is ground truth, while the former is attained from the rotation matrix $R = UV^T$ we yield from decomposing $T^Z$ into $U, S, V^T$ matrices with SVD.}
    \label{fig:tz-rotations}%
\end{figure}

For rotations on MNIST, it is trivial to parse the angle from $T^X$ as the arctan of the first column in the corresponding affine linear matrix. This is not true of $T^Z$ because it is an unconstrained prediction with entries violating the conditions of a rotation matrix. This observation is supported by Figure~\ref{fig:transform-kdeplot}, which shows the density plot for the top left entry of the $2 \times 3$ matrix of transformations. It is apparent that $T^Z$, contrary to $T^X$, cannot correspond to only translation and rotation in how it manipulates the \textit{feature} space because these entries have too high of a magnitude. Consequently, we use SVD~\cite{10.1007/BF02163027} to decompose $T^Z = USV^T$, and yield the best approximating rotation matrix $R = UV^T$~\cite{SVD_rotation_proof_paper}. We then use the angle from $R$ as the prediction and compare it the angle attained from $T^X$. Their exceptionally linear relationship can be observed for both digits in Figure~\ref{fig:tz-rotations}.

The relationship seen in these figures is further supported by a linear regression from $T^X$ to $T^Z$. We flatten both to $6$ dimensional representations and use sklearn's LinearRegression package~\cite{scikit-learn}. As expected, the $r^2$ score for each digit is exceptionally high, $0.992$ for digit 5 and $0.993$ for digit 8.

Altogether, this analysis suggests that the learned $T^Z$ has a strong linear relationship with the ground truth $T^X$, even though $T^Z$ had no prior knowledge of $T^X$. Furthermore, it also suggests that even though $T^Z$ and $T^X$ have very different statistics (exemplified by the two plots in Figure~\ref{fig:transform-kdeplot}), $T^Z \circ f_c(x_c)$ is not actually that different from $T^X$ in how it manipulates the \textit{image} space, which is what we desire.

%% file: core/2_related_work.tex
\section{Related Work}
\label{sec:related-work}

Both the problem of background and foreground separation, as well as that of manipulable synthesis, are longstanding concerns in unsupervised learning~\cite{leroux2011learning,786972,articlegreedy,Rubinstein13Unsupervised,DBLP:journals/corr/VillegasYHLL17,NIPS2017_2d2ca7ee}. As far as we know, we are the first to perform both without strong supervision such that we can independently manipulate each of the animation, the background, and the character.

Besides Dupont et al.~\cite{dupont2020equivariant}, the most closely related papers are Worrall et al.~\cite{DBLP:journals/corr/abs-1710-07307}, Olszewski et al.~\cite{DBLP:journals/corr/abs-1904-06458}, and Reed et al.~\cite{NIPS2015_5845}. The first two also use equivariance to learn representations capable of manipulating scenes. However, they do not delineate characters and backgrounds, nor do they learn the $T^Z$ from data. Dupont et al. assumes that $T^Z$ is given during training; In Worrall et al., $T^Z$ is a block diagonal composition of (given) domain-specific transformations; Olszewski et al. uses a user-provided transformation. Because we learn $T^Z$ from data, we can use datasets without ground truth like Sprites~\cite{bart_2012}. Reed et al. similarly applies transformations to frames to yield an analogous frame. Where we use affine transform, they use addition. They also require three frames as input to $T^Z$ at inference and four frames during training. Most problematic is that they require the full analogy during training. This means they cannot learn the transformation from random pair frames like we do but instead require a carefully built training set.

The papers from Siarohin et al.~\cite{siarohin2020order,siarohin2020motionsupervised} also use just pairs of images at training time and can then animate an unseen character at inference. They do not handle backgrounds independently and consequently are not able to mix and match foreground and background without artifacts. MonkeyNet~\cite{DBLP:journals/corr/abs-1812-08861} and Zablotskaia et al.~\cite{DBLP:journals/corr/abs-1910-09139} are also in this lineage and use the inductive bias provided by DensePose (and a warping module for the latter). They similarly focus on just modeling the character and place the background as is.
 
There are approaches that decompose the latent spaces with GANs~\cite{DBLP:journals/corr/MathieuZSRL16,Tulyakov_2018_CVPR} or VAEs~\cite{visualdynamics16,Kosiorek2018sqair}. Most have the same concern synthesizing the background and foreground independently with no mechanism for delineating the two, even though there may be mechanisms for parsing the foreground like in Kosiorek et al.~\cite{Kosiorek2018sqair}. Vondrick et al.~\cite{DBLP:journals/corr/VondrickPT16} does have such a mechanism, however it serves only to separate background from character pixels through a masking function. Because our mechanism acts on the learned character function, we are able to model the animation as well.



%% file: core/7_conclusion.tex
\section{Conclusion}

We have presented a self-supervised framework for learning an equivariant renderer capable of delineating the background, the characters, and their animation such that it can manipulate and synthesize each independently. Our framework requires only video sequences. We tested it on two datasets chosen to highlight the model's capability in handling both truly affine movements (Moving MNIST) as well as movements from video (Sprites). We then showed that it can produce convincing and consistent scene manipulations, both of background and animation transfer. 
Our analysis revealed a strong association between the learned $T^Z$ and the ground-truth $T^X$, which was never seen during training, implying the effectiveness of the proposed self-supervised learning criterion. We also observed aspects of $T^Z$ not fully explained by $T^X$.

An assumption that $T^X$ is affine does not hold in general; a clear counterexample is videos with nonlinear lighting effects. However, there is no reason why such an assumption on $T^Z$ could not hold. Consequently, we do not see any barriers to our approach working on real-world examples such as the motivating stop-motion animation. While our results are not yet strong enough on complex real-world applications (see Section~\ref{sec:fashion}), we see this work as being a promising step in that direction and leave that advance for future work.


%% file: core/8_broader_impact.tex

%% file: appendix.tex
\section*{Appendix}

\begin{figure}[h]
    \centering
    \includegraphics[width=0.41\textwidth]{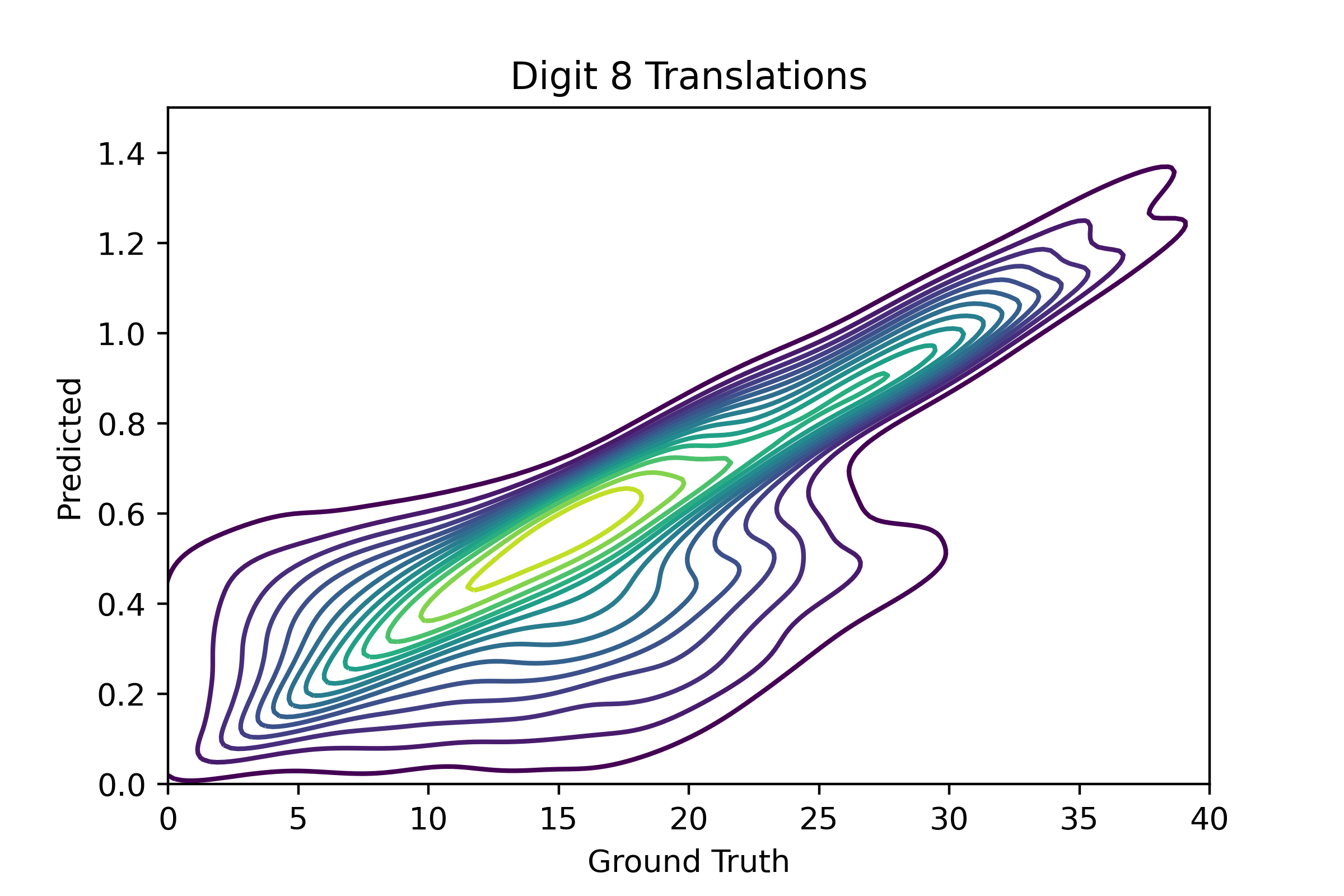}
    \caption{Matching KDE plot for digit 8. See Figure~\ref{fig:tz-translations} in the main text for digit 5.}
    \label{fig:tz-translations8}%
\end{figure}

\begin{figure}[h]
    \centering
    \includegraphics[width=0.41\textwidth]{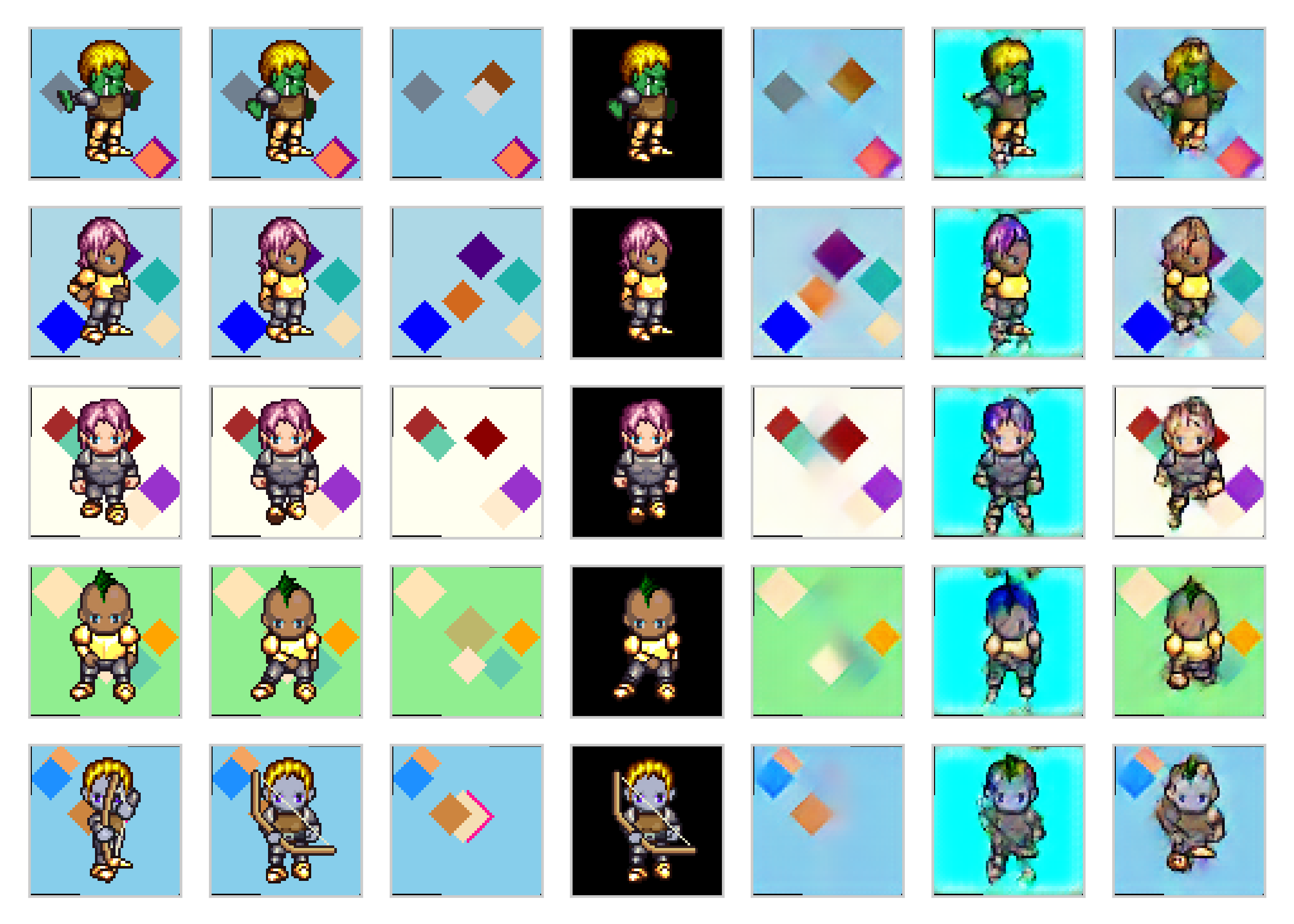}
    \caption{Panel referenced by Figure~\ref{fig:zerooutboth} showing five results from a model trained with sprites undergoing x-translation during training in addition to the normal changes in the character animation. Observe that the model learns a clean separation of background and character in this case.}
    \label{fig:reconboth-212}%
\end{figure}

\begin{figure}[h]
    \centering
    \includegraphics[width=0.41\textwidth]{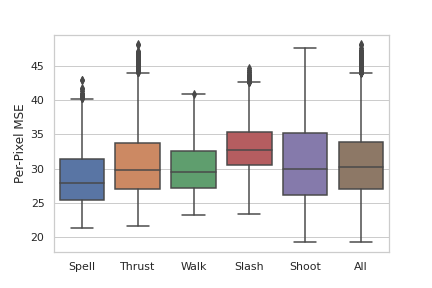}
    \caption{MSE plot for reconstructing analogies on blank canvases using a model trained with backgrounds. See Figures~\ref{fig:pixelmse},~\ref{fig:analogyrecon} for matching analysis and Figure~\ref{fig:recon-removeback-examples} for examples. Note that we did not use black backgrounds for the reconstruction but gray ones instead. This is because the model was never trained with the color black and it would be out of scope to generalize to a completely new color.}
    \label{fig:recon-removeback-mse}%
\end{figure}

\begin{figure}[h]
    \centering
    \includegraphics[width=0.41\textwidth]{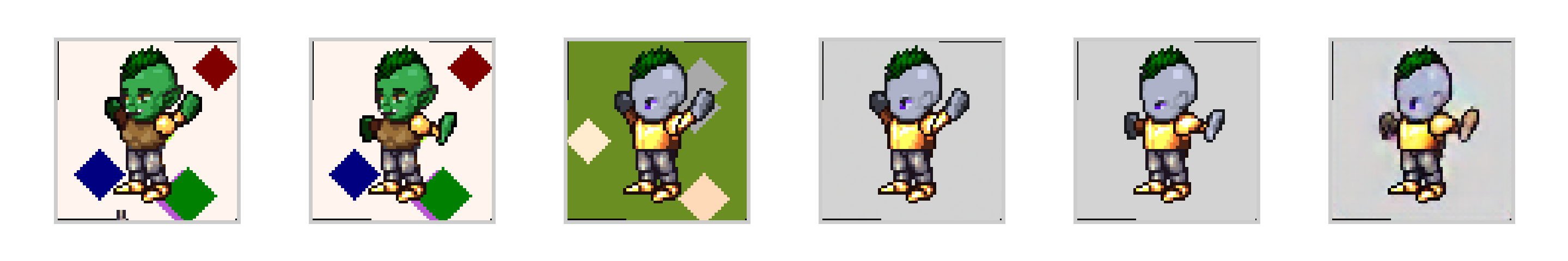}
    \includegraphics[width=0.41\textwidth]{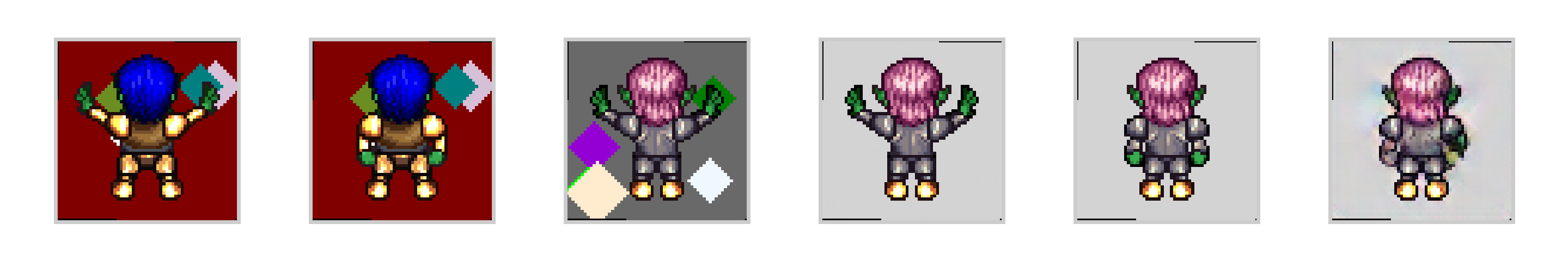}
    \caption{Example sequences of removing the background. The first two columns are the animation, the third column is the character, and the fourth column is the character on a bare background. These inputs yield the output seen in the sixth column which can be compared to the ground truth expected output in the fifth column.}
    \label{fig:recon-removeback-examples}%
\end{figure}

\begin{figure*}[ht]
    \centering
    \includegraphics[width=.3\textwidth]{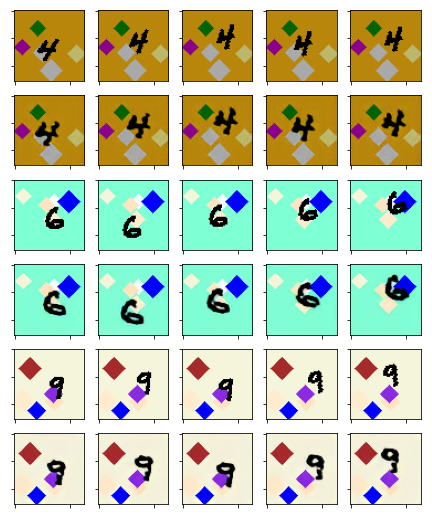}
    \caption{\textbf{Reconstructions}: The first, third, and fifth rows are original sequences. The second, fourth, and sixth rows are reconstructions of the prior row where $T^Z$ is fixed as the same transformation as $T^X$. Note that in these scenarios, the results are not as strong as when $T^Z$ is a learned function.}
    \label{fig:reconstruct}
\end{figure*}

\begin{figure*}[ht]
    \centering
    \includegraphics[width=.3\textwidth]{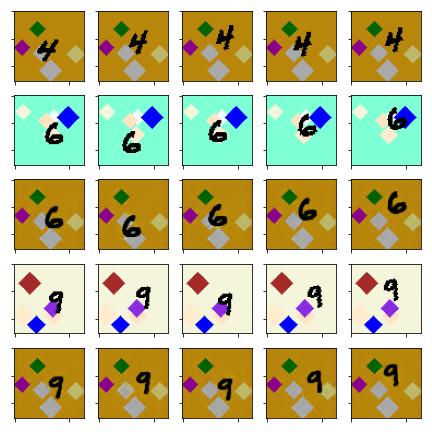}
    \caption{\textbf{Backgrounds}: The first, second, and fourth rows are originals. The third and fifth rows are the prior row but with the background changed to that of the first sequence.}
    \label{fig:background}
\end{figure*}

\begin{figure*}[ht]
    \centering
    \includegraphics[width=.3\textwidth]{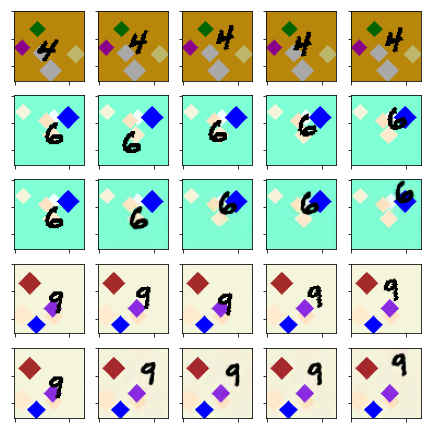}
    \caption{\textbf{Transformations}: The first, second, and fourth rows are originals. The third and fifth rows are the prior row but the transformation is a function of the first row.}
    \label{fig:background}
\end{figure*}

\begin{figure*}[!ht]
    \centering
    \subfloat[\centering Analogy MSE]{{\includegraphics[width=9cm]{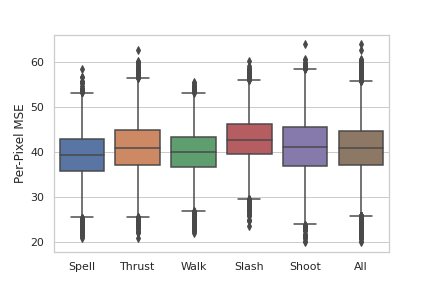} }}%
    \qquad
    \subfloat[\centering Background-only MSE]{{\includegraphics[width=9cm]{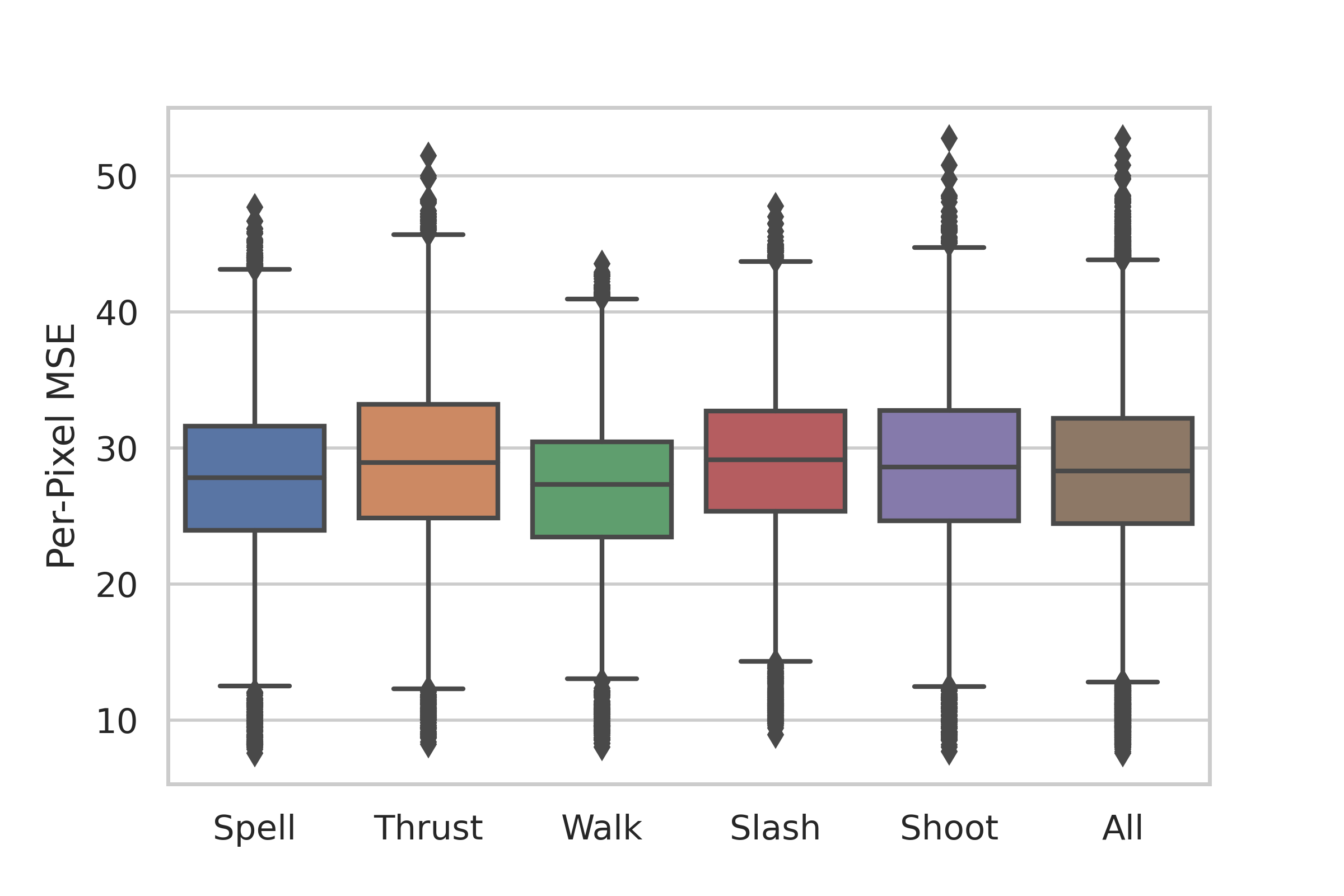} }}%
    \qquad
    \subfloat[\centering Character-only MSE]{{\includegraphics[width=9cm]{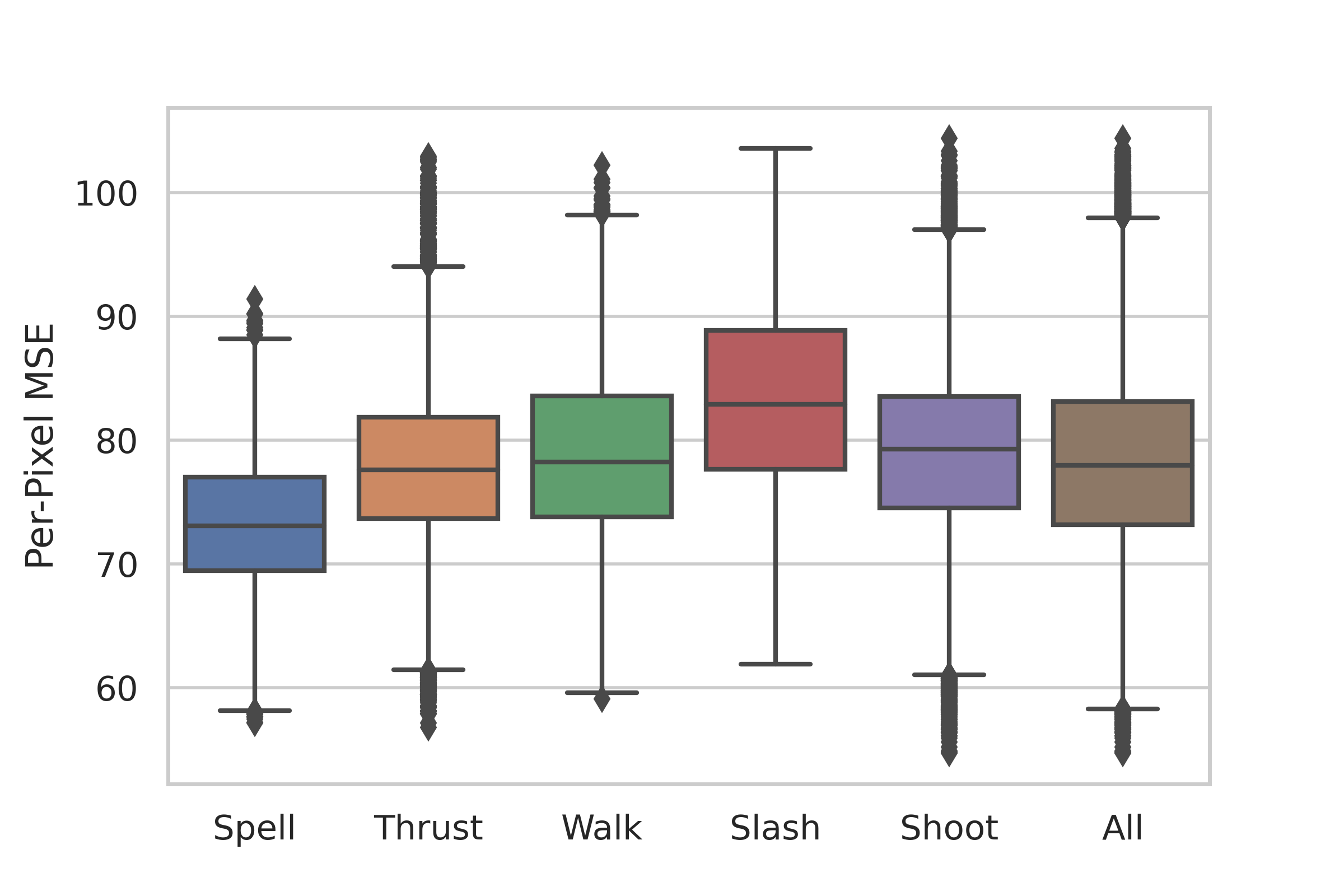} }}%
    \qquad
    \caption{Analogy reconstruction with backgrounds. As expected, the model's MSE increases when incorporating backgrounds. Panels (b) and (c) show that this is dominantly due to reconstructing the character and not the background.}
    \label{fig:sprite-color-recon-breakdown}%
\end{figure*}

\begin{figure}[!ht]
    \centering
    \includegraphics[width=0.33\textwidth]{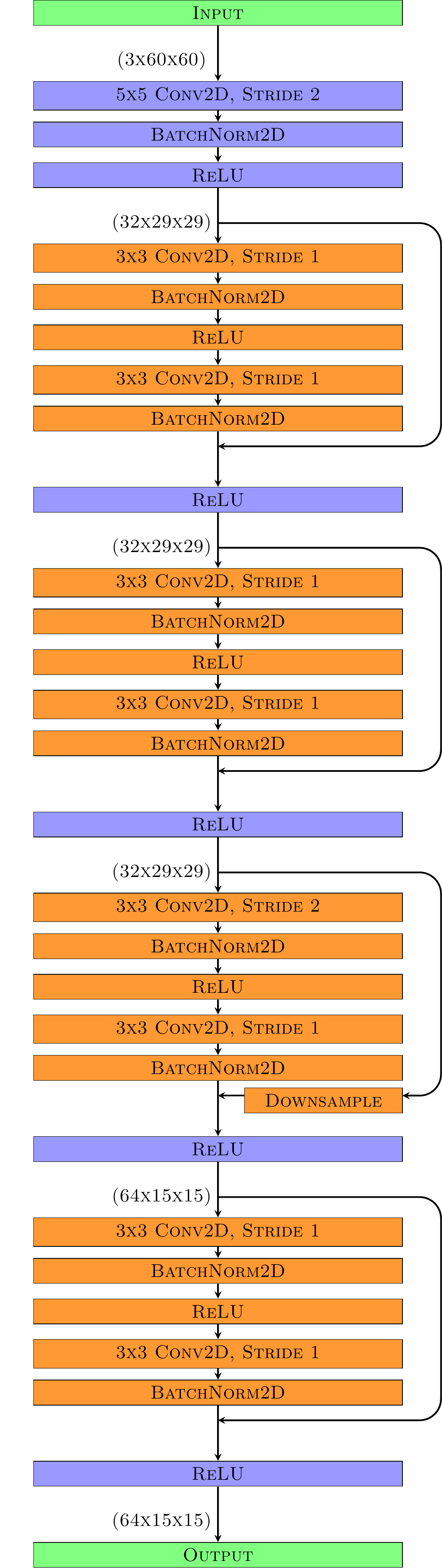}
    \caption{Architecture of the Encoder for both characters and backgrounds.}
    \label{fig:encoder_architecture}
\end{figure}

\begin{figure}[!ht]
    \centering
    \includegraphics[width=0.33\textwidth]{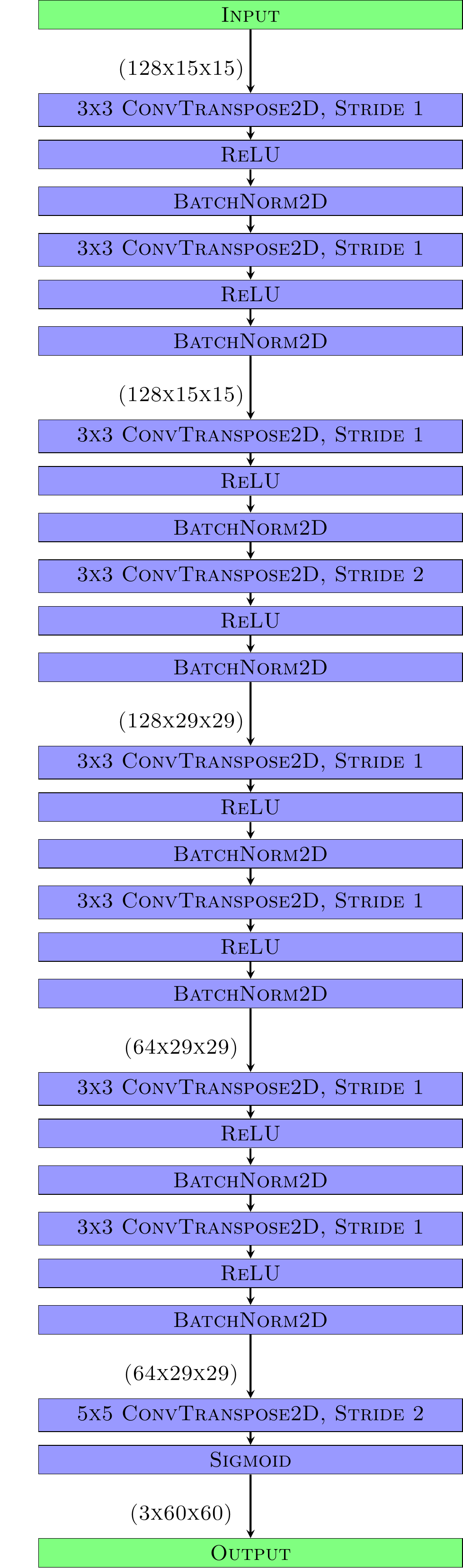}
    \caption{Architecture of the Decoder.}
    \label{fig:decoder_architecture}
\end{figure}

\begin{figure*}[!ht]
    \centering
    \includegraphics[width=0.66\textwidth]{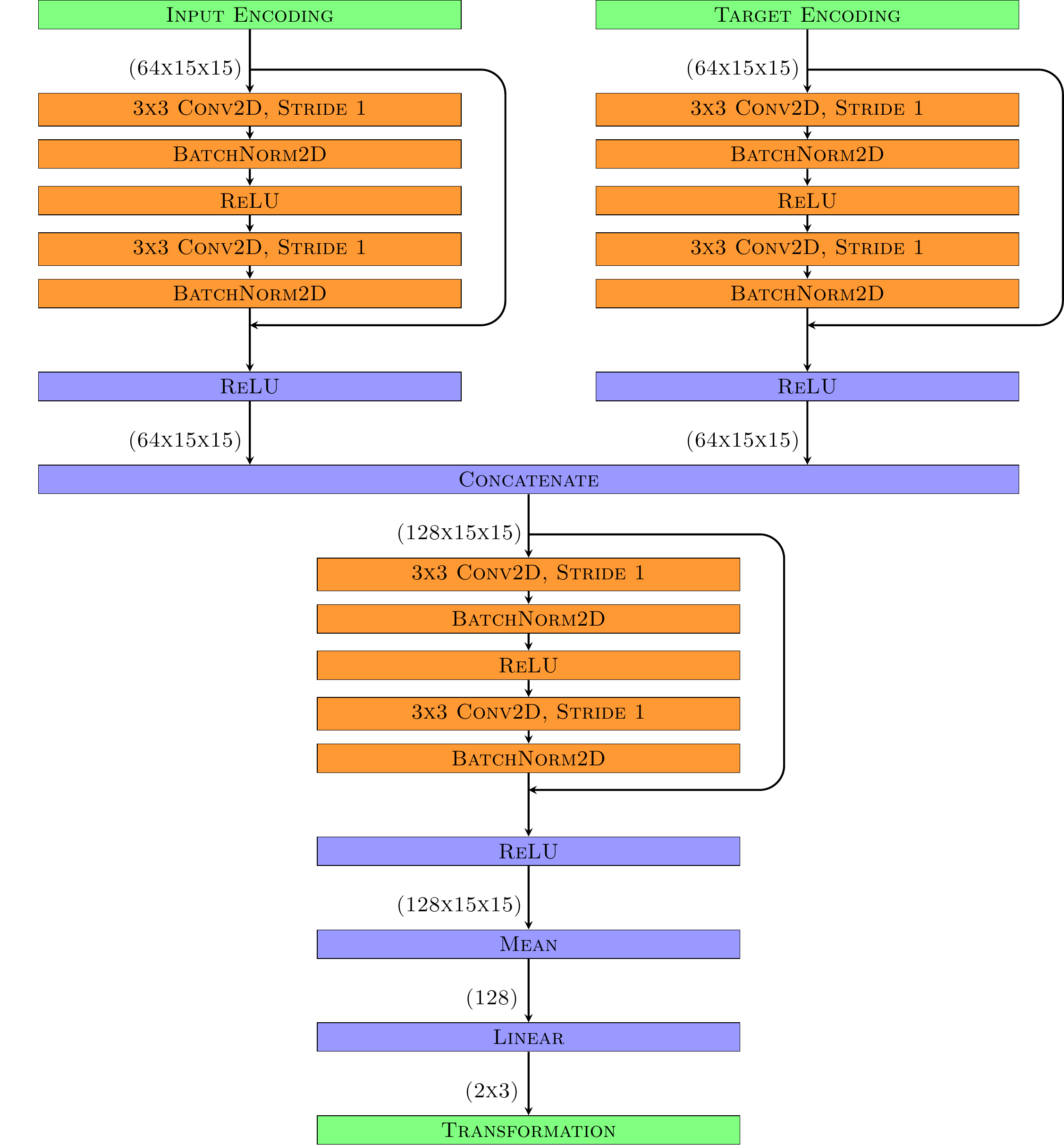}
    \caption{Architecture of $T^Z$.}
    \label{fig:siamese_architecture}
\end{figure*}

\begin{figure*}[!ht]
    \centering
    \includegraphics[width=0.66\textwidth]{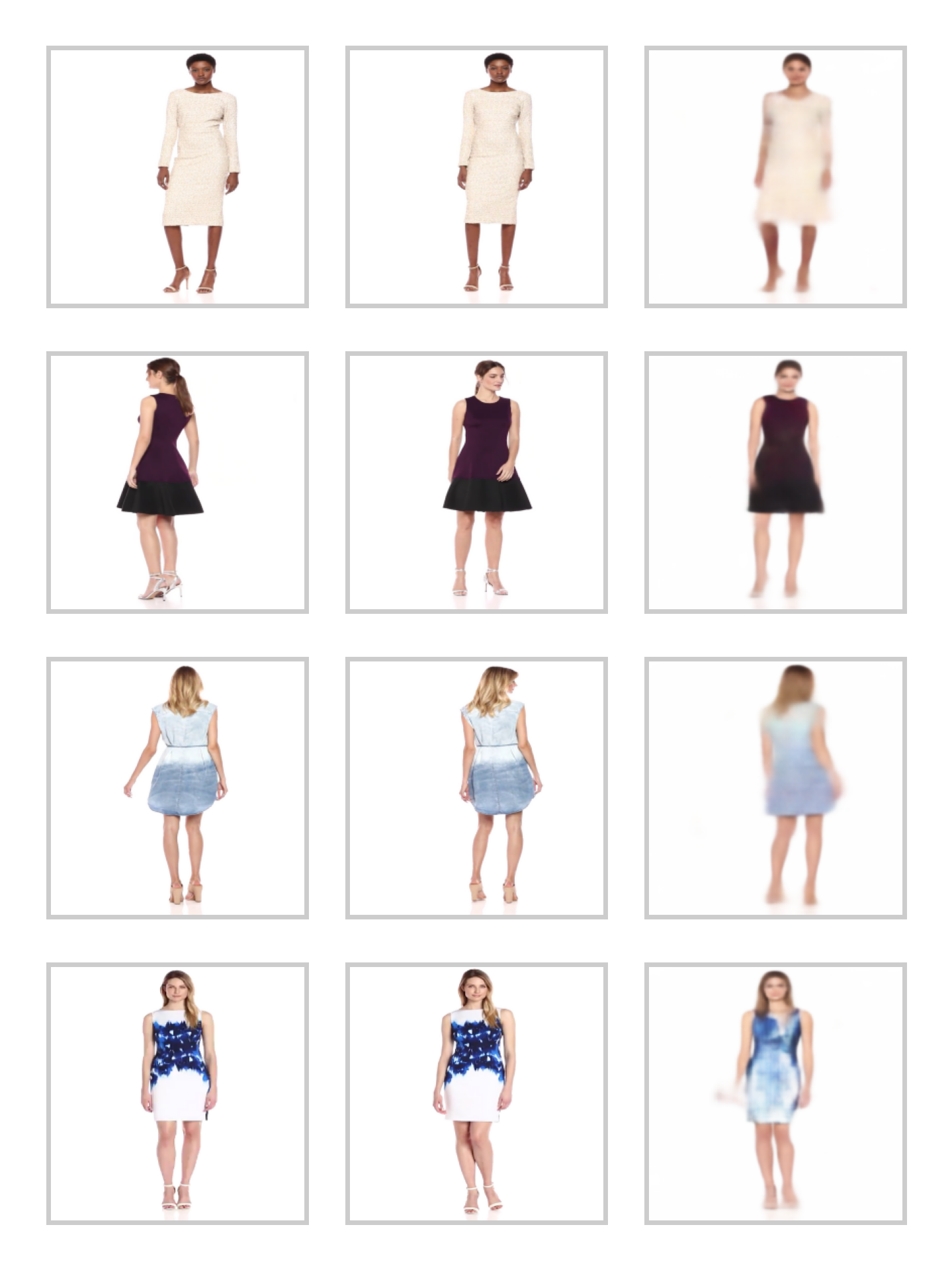}
    \caption{Fashion reconstructions. The first column serves as $x_c$, $x_{t1}$, and $x_b$, the second as $x_{t2}$ and the target. In the first row, we can see that the model is not sure what to do with the legs. In the second and third rows, the model has successfully repositioned the character but remains blurry.}
    \label{fig:fashion-recon}
\end{figure*}

\begin{figure*}[!ht]
    \centering
    \includegraphics[width=15cm]{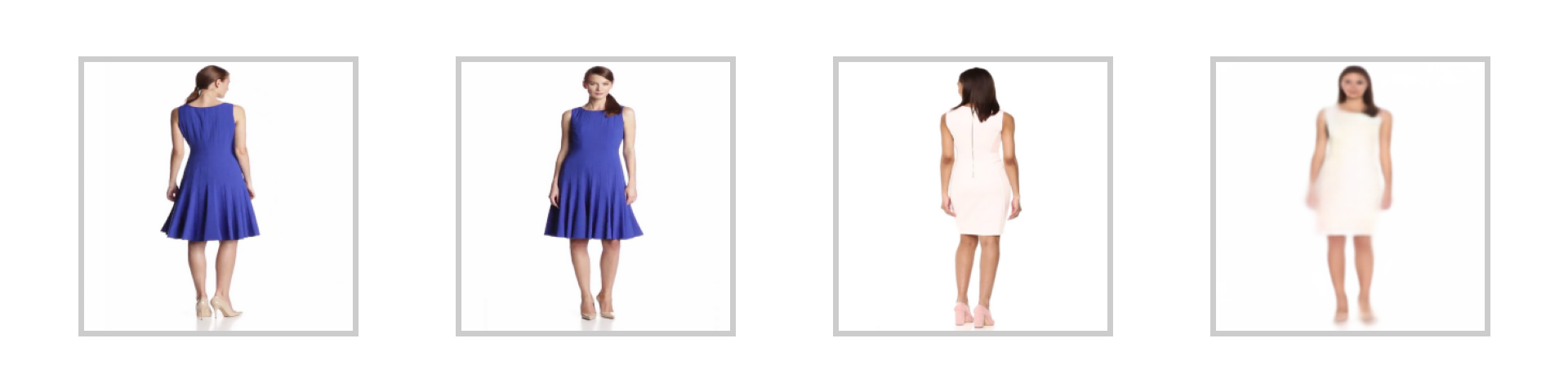}
    \includegraphics[width=15cm]{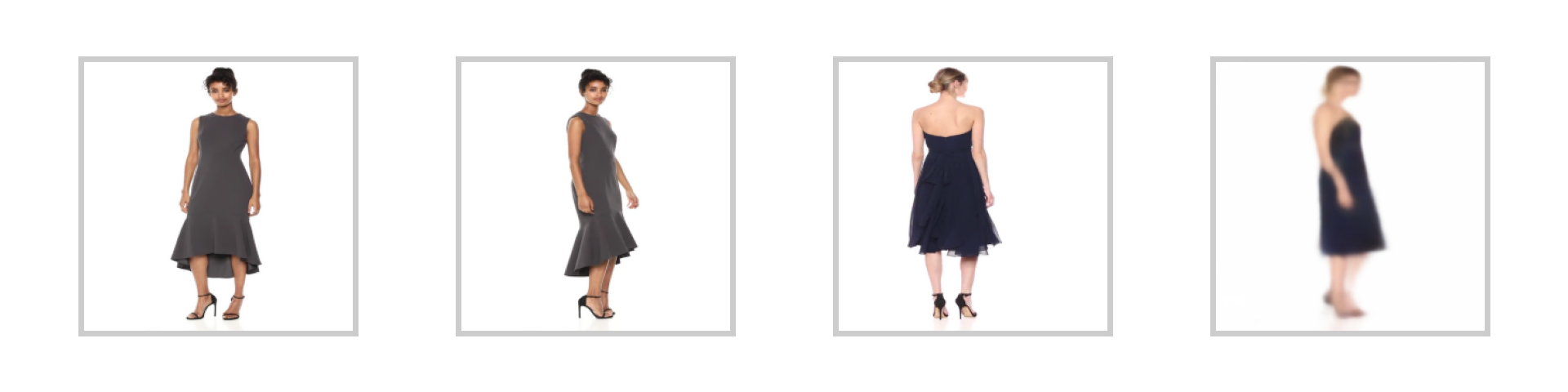}
    \includegraphics[width=15cm]{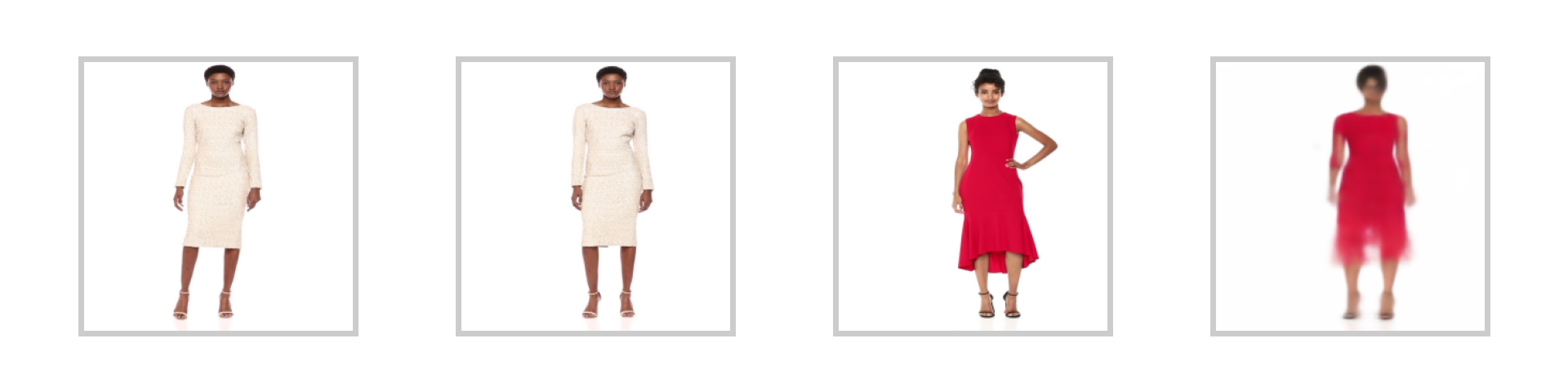}
    \includegraphics[width=15cm]{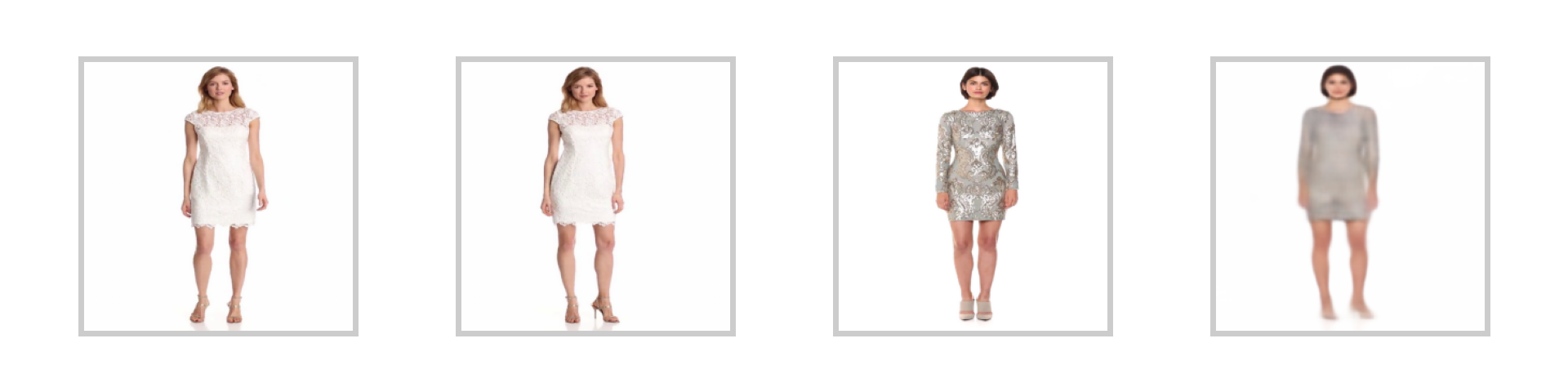}
    \caption{Fashion analogies. The first and second columns represent the animation to be done to the character in the third column. The first and fourth rows show successful animations, even with the slight animation in the latter. The second row is facing the wrong direction. The third row should have her arm still up and her leg moved to the side.}
    \label{fig:fashion-analogy}%
\end{figure*}